\documentclass[10pt]{article}
\usepackage[preprint]{tmlr}

\usepackage{algorithm}
\usepackage{algorithmicx}
\usepackage{algpseudocode}
\usepackage{amsmath,amsfonts}
\usepackage{booktabs}
\usepackage{caption}
\usepackage{hyperref}
\usepackage{cleveref}
\usepackage{enumitem}
\usepackage{dblfloatfix}
\usepackage{epsfig}
\usepackage{graphicx}
\usepackage[utf8]{inputenc}
\usepackage{multirow}
\usepackage{subcaption}
\usepackage{tabularx}
\usepackage{textcomp}
\usepackage{times}
\usepackage{wrapfig}
\usepackage{xcolor}
\usepackage{xspace}
\usepackage{soul}
\usepackage{url}
\setstcolor{red}

\newcommand{\adv}{$\mathcal{A}$\xspace}
\newcommand{\victim}{$\mathcal{V}$\xspace}
\newcommand{\victimfnc}{$F_{\mathcal{V}}$\xspace}
\newcommand{\advfnc}{$F_{\mathcal{A}}$\xspace}

\newcommand{\mtop}{Monet-to-Photo\xspace}
\newcommand{\stoa}{Selfie-to-Anime\xspace}
\newcommand{\superres}{Super-Resolution\xspace}

\title{Good Artists Copy, Great Artists Steal: Model Extraction Attacks Against Image Translation Models}

\author{\name Sebastian Szyller \email contact@sebszyller.com \\
      \addr Aalto University
      \AND
      \name Vasisht Duddu \email vasisht.duddu@uwaterloo.ca \\
      \addr University of Waterloo
      \AND
      \name Tommi Buder-Gröndahl \email tommi.grondahl@helsinki.fi\\
      \addr University of Helsinki
      \AND
      \name N. Asokan \email asokan@acm.org \\
      \addr University of Waterloo \& Aalto University}

\begin{document}

\maketitle


\begin{abstract}
Machine learning models are typically made available to potential client users via inference APIs. \emph{Model extraction attacks} occur when a malicious client uses information gleaned from queries to the inference API of
a victim model \victimfnc to build a \emph{surrogate model} \advfnc with comparable functionality. Recent research has shown successful model extraction of image classification, and natural language processing models.

In this paper, we show the first model extraction attack against real-world generative adversarial network (GAN) \emph{image translation models}. We present a framework for conducting such attacks, and show that an adversary can successfully extract functional surrogate models by querying \victimfnc 
using data from the same domain as the training data for \victimfnc. The adversary need not know \victimfnc's architecture or any other information about it beyond its intended task. 

We evaluate the effectiveness of our attacks using three different instances of two popular categories of image translation: (1) \stoa and (2) \mtop (image style transfer), and (3) \superres (super resolution). Using standard performance metrics for GANs, we show that our attacks are effective. 
Furthermore, we conducted a large scale (125 participants) user study on \stoa and \mtop to show that human perception of the images produced by \victimfnc and \advfnc can be considered equivalent, within an equivalence bound of Cohen's $d=0.3$.

Finally, we show that existing defenses against model extraction attacks (watermarking, adversarial examples, poisoning) do not extend to image translation models.
\end{abstract}


\section{Introduction}\label{introduction}

Machine learning (ML) models have become increasingly popular across a broad variety of application domains.
They range from tasks like image classification and language understanding to those with strict safety requirements like autonomous driving or medical diagnosis.
Machine learning is a multi-billion dollar industry~\citep{techworld:2018} supported by technology giants, such as Microsoft, Google, and Facebook.

\begin{figure}[t]
    \centering
    \resizebox{1.\columnwidth}!{
    \begin{tabular}{ccc}
        \includegraphics[width=0.25\columnwidth]{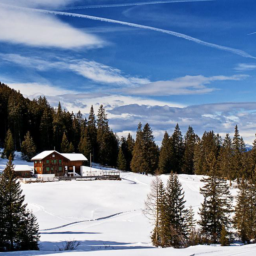} &
        \includegraphics[width=0.25\columnwidth]{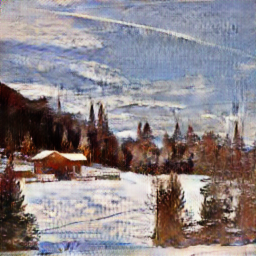} &
        \includegraphics[width=0.25\columnwidth]{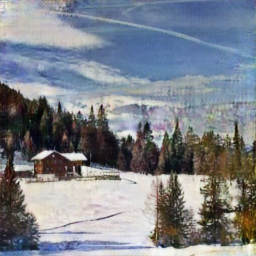} \\
        \includegraphics[width=0.25\columnwidth]{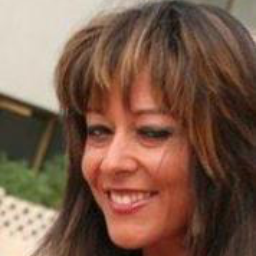} &
        \includegraphics[width=0.25\columnwidth]{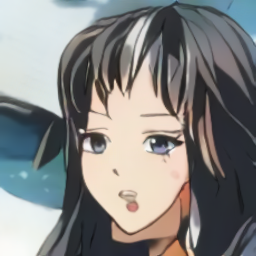} &
        \includegraphics[width=0.25\columnwidth]{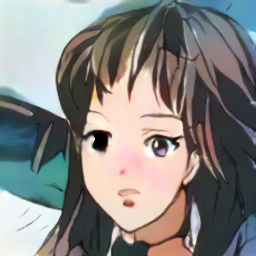} \\
    \end{tabular}}
    \caption{Example comparison of style transfer. \textbf{left}: original, \textbf{center}: \victimfnc output, \textbf{right}: \advfnc output. Top row: \mtop style transfer, bottom row: \stoa style transfer. Images are down-scaled to fit the page.}
    \label{fig:example_comparison1}
\end{figure}

\begin{figure*}[t]
    \centering
    \resizebox{\columnwidth}!{
    \includegraphics[]{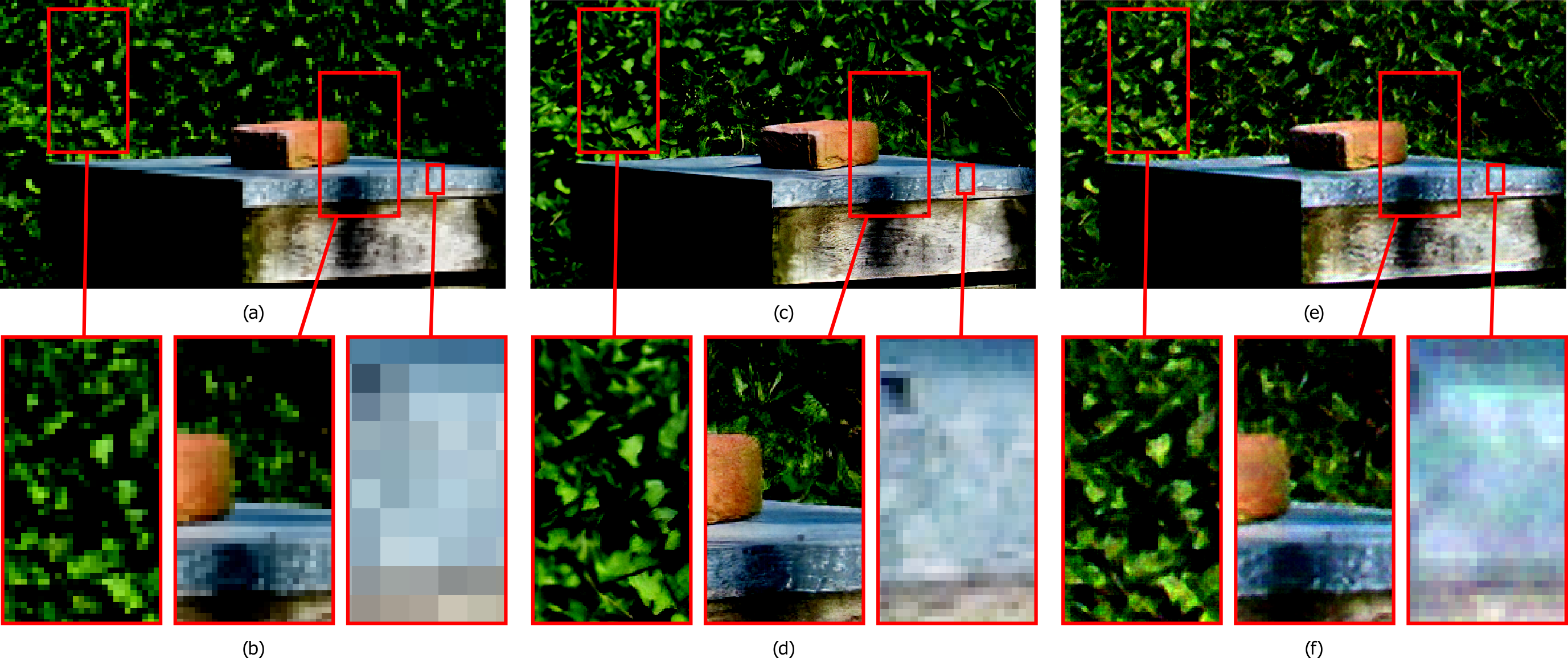}}
    \caption{Example comparison of a \superres transformation. \textbf{Left}: original (a), \textbf{center}: \victimfnc output (c), \textbf{right}: \advfnc output (e). Image up-scaled x4 from 175x100. We zoom in selected parts of the images (b,d,f) to show the difference in resolution among the three images.}
    \label{fig:example_comparison2}
\end{figure*}

Recently \emph{image translation} applications have become popular in social media.
Examples include coloring old photos~\citep{pix2pix}, applying cartoon based filters~\citep{selfie2animeService} or generating fake images of people (DeepFakes~\citep{tolosana2020deepfakes}).
Features like face filters or face transformations are now an integral part of various popular applications such as TikTok, Snapchat or FaceApp.
Such features are costly to create.
They require data collection and sanitization, engineering expertise as well as computation resources to train \emph{generative adversarial network} (GAN) models to implement these features.
This represents a high barrier-to-entry for newcomers who want to offer similar features.
Hence, effective image translation models confer a business advantage on their owners.

Use of ML models is typically done in a \emph{black-box} fashion -- clients access the models via inference APIs, or the models are encapsulated within on-device sandboxed apps.
However, a malicious client (adversary) can mount \emph{model extraction attacks}~\citep{tramer2016stealing} by querying a model and using the responses to train a local \emph{surrogate model} \advfnc that duplicates the functionality of the victim model \victimfnc.
So far model extraction attacks have been shown to be successful in stealing image classifiers~\citep{tramer2016stealing, papernot2017practical, juuti2019prada,orekondy2018knockoff, correia2018copycat, jagielski2020highfidelity}, and natural language processing models~\citep{krishna2020thieves,wallace2021translation}.

In this paper,
we show that an adversary can successfully build surrogate models mimicking the functionality of several types of image translation models (Figures~\ref{fig:example_comparison1}-\ref{fig:example_comparison2}).
Our attack does not require the adversary to have access to victim's training data, knowledge of victim model parameters or its architecture.
We make the following contributions:
\begin{enumerate}
    \item present the \textbf{first model extraction attack against image translation models} (Section~\ref{approach}), and
    \item empirically demonstrate its effectiveness via:
    \textbf{1) quantitative analyses} using perceptual similarity metrics on three datasets spanning two types of image translation tasks: style transfer and image super resolution; the differences between the victim and surrogate models, compared to the target, are in the following ranges -- \stoa: Frechet Information Distance (FID) $13.36-39.66$, \mtop\footnote{The task is translating from photos to Monet paintings, but we refer to it as ``\mtop'' to be consistent with the name of the dataset~\citep{cyclegan}.}: FID $3.57-4.40$, and \superres: Structural Similarity: $0.06-0.08$ and Peak Signal-to-Noise Ratio: $1.43-4.46$ (Section~\ref{evaluation});
    \textbf{2) an extensive user study} ($n=125$) on the style transfer tasks (\stoa and \mtop) showing that human perception of images produced by the victim and surrogate models are similar, within an equivalence bound of Cohen's $d=0.3$. (Section~\ref{userstudy}); and
    \item highlight the need for novel defenses by showing that \textbf{existing defenses against model extraction are ineffective} for image translation models (Section~\ref{defense}).
\end{enumerate}


\section{Background and Related Work}\label{background}

In this section, we introduce the background necessary to understand the rest of this paper.
Also, we give an overview of the use of generative adversarial networks (GANs) for image translation tasks, and a set of standardized metrics used to measure the effectiveness of such translations.

\subsection{Model Extraction Attacks and Defenses}
The goal of the adversary \adv in model extraction attacks is to build a surrogate model to mimic the functionality of a victim model~\citep{orekondy2018knockoff, papernot2017practical}. The motivations for model extraction can be manifold: to use the surrogate model to offer a competing service~\citep{correia2018copycat,orekondy2018knockoff}, or to probe the surrogate model to find adversarial examples that transfer to the victim model itself~\citep{juuti2019prada, papernot2017practical}.
\adv conducts the attack by repeatedly querying the inference interface of the victim and using the responses to train the surrogate.

The techniques used for model extraction can differ depending on the assumptions regarding the knowledge and capabilities of the adversary. It is typical to assume that \adv does not have access to the victim's training data but may know the general domain (e.g., pictures of animals)~\citep{papernot2017practical, correia2018copycat}.
It is also customary to assume that \adv does not know the exact architecture of the victim model.
However, \adv may infer information about the victim architecture using side-channel attacks~\citep{batina2019elecmagsidechannels,sidechannel,timingsc}, snooping on shared caches~\citep{yan2020sharedresources}, or matching model's responses to a series of queries with known architectures~\citep{joon2018towards}.

Previous work on model extraction attacks focuses primarily on image classification tasks~\citep{tramer2016stealing,papernot2017practical,juuti2019prada}.
Examples include attacks against ImageNet-scale models where \adv's goal was \emph{functionality stealing}~\citep{orekondy2018knockoff,correia2018copycat,jagielski2020highfidelity}.
Other than image classification, there is a large body of work targeting other types of models - NLP models~\citep{krishna2020thieves,wallace2021translation}, GNNs~\citep{he2020stealinglinks}, and RNNs~\citep{takemura2020rnns}.
Existing attack against GANs~\citep{hu2021ganstealing} is limited to models that generate images from latent codes, and is not applicable to image translation models.

Existing model extraction defenses either attempt to identify \adv by studying the query distribution~\citep{juuti2019prada,boogeyman} or try to identify queries that are close to decision boundaries~\citep{quiring2018forgottensib,zheng2019bdpl}, or explore an abnormally large part of the hyperspace~\citep{Kesarwani2017model}.
Alternatively, the defender can perturb the prediction vector before sending it back to the client to slow down the construction of effective surrogate models~\citep{kariyappa2019adaptivemisinformation,orekondy2020predictionpoisoning}.

Instead of detecting or defending against model extraction attacks, the defender can watermark their model and try to claim ownership if \adv makes their surrogate model public.
Watermarks can be embedded into the model during training by modifying its loss function~\citep{jia2021entangled} that makes it more likely that the watermarks transfer to the surrogate model.
Alternately, model responses can be modified at the API level to embed a client-specific watermark into \adv's training data~\citep{szyller2020dawn}.
Both schemes are based on the idea that the watermark must extend to \emph{all} models \emph{derived from} \victimfnc, and not just the \victimfnc. Watermarking schemes that do not consider this were shown not be resilient to model extraction attacks~\citep{shafieinejad2019stealingvswatermarking}.

While there do exist watermarking schemes for GANs, they either primarily assume a black-box generator style of GAN~\citep{yu2021artificial} or that \adv is going to use specific data to launch the attack~\citep{zhang2020model}.

\subsection{Generative Adversarial Networks}

Generative Adversarial Networks (GANs) are generative models which learn the underlying data distribution as a game between two models: a generator model $G$ which generates new images from input noise and a discriminator model $D$ which attempts to distinguish real images from the training data and fake images from $G$~\citep{gan}.
This is modeled as a minimax optimization problem: $\min_{q_{\phi}} \max_{D} V(G, D)$, where $D$ maximises its gain $V$ obtained by correctly distinguishing real and fake generated images.
$G$ minimizes the maximum gain by $D$ and iteratively improves the quality of images to be close to the original training data points while fooling $D$.
The overall optimization objective can be formulated as follows:

{\small
\begin{align}\label{ganeq}
\nonumber V(G, D) &= \mathbb{E}_{x \sim p_{data}(x)}[log(D(x))] \\
                 &+ \mathbb{E}_{z \sim p_{z}(z)}[1-log(D(G(z)))]
\end{align}}

where $p_{data}$ is the distribution of the training data and $p_{z}$ is the distribution learnt by $G$.
This optimization is solved by alternatively training $D$ and $G$ using some variant of gradient descent, e.g. SGD or Adam.
The outstanding performance of GANs have resulted in their widespread adoption for different applications addressing several image processing and image translation tasks.
In this work, we consider two main applications of image to image translations: style transfer~\citep{cyclegan,pix2pix,ugatit} and image super resolution~\citep{srgan}.

\noindent\textbf{Style Transfer.}
GANs have been used to translate images from one style to another~\citep{pix2pix}.
Prior work explored several image style transfer tasks such as changing seasons, day to night, black and white to colored, and aerial street images to maps~\citep{pix2pix,cyclegan,ugatit}.
Some specific tasks have been extensively adopted as filters in social networks such as Instagram, Snapchat and FaceApp, e.g. changing human faces to anime characters~\citep{ugatit}, aging or swapping genders~\citep{stylegan,stargan}.
These are trained using GANs of different types.
Style transfer GANs can be trained using paired image-to-image data with a Pix2Pix model~\citep{pix2pix} in a supervised fashion - each unstyled image has its styled counterpart.
For complex tasks, such supervised image to image translation training data is difficult, if not impossible, to obtain.
To address this, training GANs using unpaired and uncorrelated mappings of the images in the input source domain to the target output domain has been proposed (e.g. CycleGAN, UGATIT~\citep{ugatit,cyclegan}).

\noindent\textit{Pix2Pix}~\citep{pix2pix}
architecture uses a UNet Generator Model which downsamples an input image from the source domain and outputs an upsampled image in the target domain.
It uses a PatchGAN discriminator network which takes the input image from the source domain and the image from the target domain to determine whether the generated output image from the target domain is from the real distribution or from the UNet.
To train a Pix2Pix model, the UNet generator is trained to minimize the loss of correctly translating the input source domain image to the target domain image ($\mathit{IdentityLoss}$) while additionally minimizing the adversarial loss that captures the success of discriminator, to produce realistic target domain images.
The overall loss optimized includes: $\mathit{GeneratorLoss} = \mathit{AdversarialLoss} + \lambda * \mathit{IdentityLoss}$.

\noindent\textit{CycleGAN}~\citep{cyclegan}
consists of two GANs.
One of the GANs maps images from the source domain to the target domain while the second GAN maps images from the target domain back to the source domain.
Both generators are similar to generators in Pix2Pix and map images from one domain to the other by downsampling the image, followed by upsampling the image to the target domain.
Like Pix2Pix, CycleGAN uses adversarial loss with respect to the discriminators to ensure that the generator outputs domain images close to real images, and Identity Loss to ensure similarity between the generated image and real target domain image.
Additionally, CycleGAN uses the optimized cycle consistency loss.
It ensures that an image when mapped from the source domain to the target domain and back to the source domain are as similar as possible.

\noindent\textit{UGATIT}~\citep{ugatit}
utilizes convolutional neural networks as an attention module to identify and extract major features in images, in addition to the architecture and optimization of the CycleGAN.
This allows for finer optimization while translating input images to the target domain.

\noindent\textbf{Image Super Resolution.}
Maps low resolution images to high resolution images using a GAN.
The generator includes residual blocks that alleviate the vanishing gradient problem and hence, enable deeper models that give better results.
The generator outputs a high resolution (HR) image that is an up-scaled version of its low resolution input counterpart.
The discriminator differentiates between the generated HR image with the ground truth HR image.
The generator parameters are updated similarly to a generic GAN (Equation~\ref{ganeq}).

\subsection{Metrics}\label{metrics}

We consider several metrics for comparing the quality of images generated by the victim and surrogate models.
Unlike classification models that rely on accuracy with respect to ground truth, we need to be able to objectively compare the performance of the victim and surrogate models without necessarily having access to any ground truth.

\noindent\textbf{Structural Similarity (SSIM)}~\citep{1284395}
of two image windows $x$ and $y$ of same size $N\times N$, is:

{\small
\begin{equation}
SSIM(x,y) = \frac{(2\mu_{x}\mu_{y}+c_1)(2\sigma_{xy}+c_2)}{(\mu_{x}^2 + \mu_{y}^2 +c_1)(\sigma_{x}^2 + \sigma_{y}^2 + c_2)}
\end{equation}}

where $\mu$ is the mean of pixels over the windows, $\sigma^2$ is the variance of pixel values and $\sigma_{xy}$ is the covariance, $c_1$ and $c_2$ are constants to ensure numerical stability.
SSIM compares two images with one of the images as a reference. Dissimilar images are scored as $0.0$ while the same images are given a score of $1.0$.

\noindent\textbf{Peak Signal to Noise Ratio (PSNR)}~\citep{5596999}
is the ratio of the signal (image content) and the noise that lowers the fidelity of the image:

{\small
\begin{equation}
PSNR = 20 . log_{10}\frac{MAX}{\sqrt{MSE}}
\end{equation}}

where MAX is maximum range of pixel value in the image given by $2^{b}-1$ where b is the number of bits.
MSE is the mean squared error which is the squared difference between the image pixel values and the mean value.

We use SSIM and PSNR to compare the quality of the generated super resolution image to the original HR ground truth image.
For the style transfer tasks, we consider Frechet Inception Distance instead.

\noindent\textbf{Frechet Inception Distance (FID)}
measures the distance between the features of real and generated images, and is used to estimate the quality of images from GANs~\citep{inceptionscore,ganmetric}.
FID uses the features from the last pooling layer before the output prediction layer from the Inceptionv3 network.
The distance between the feature distributions for two sets of images is computed using Wasserstein (Frechet) distance.
Low scores correspond to higher quality generated images but the magnitude of what constitutes high or low FID depends on the domain and complexity of measured images.



\section{Problem Statement}\label{sec:problem-statement}

\noindent\textbf{Adversary Model.}\label{sec:adversary-model}
The adversary \adv wants to build a surrogate model \advfnc to mimic the functionality of a model \victimfnc belonging to the victim \victim.
\adv is aware of the purpose of \victimfnc, e.g. style transfer from Monet paintings, and can choose an appropriate model architecture of \advfnc.
However, \adv is not aware of the exact architecture of \victimfnc.
\adv conducts the attack with any data $X_{\mathcal{A}}$ they want but they have no knowledge of the exact training data \victim used.
Crucially, we assume that \adv does not have access to any \emph{source style} images $S_X$ that contain the style template $S$.
We assume that \victim wants to keep $S_X$ secret because it is valuable and difficult to obtain.
\adv has only \emph{black-box} access to \victimfnc.
\adv uses $X_{\mathcal{A}}$ and $X_{\mathcal{A}_S} = F_\mathcal{V}(X_\mathcal{A})$ to train \advfnc in any way they want.
\adv's goal is to duplicate the \emph{functionality} of \victimfnc as closely as possible.

\noindent\textbf{Goals.}\label{sec:goals}
\adv's goal is not to optimize the style transfer task itself but rather to achieve a level of performance comparable to \victimfnc.
Therefore, they use \victimfnc as an oracle to obtain its \emph{``ground truth''} for training \advfnc.
\adv wants to either keep using the service without paying or to offer a competitive service at a discounted price.
\adv is not trying to obtain the exact replica of \victimfnc.
Upon a successful model extraction attack, images styled by \advfnc should be comparable to those produced by \victimfnc.
Assuming \victimfnc is a high quality style-transfer model, then the attack is considered successful if $F_\mathcal{V}(X) \approx F_\mathcal{A}(X)$.


\section{Extracting Image Translation Models}\label{approach}

We target two most popular applications of GANs for image translation: image style transfer (\stoa and \mtop) and image super resolution.
The attack setting considers \victim's model \victimfnc: $X \rightarrow X_S$ that maps an unstyled original image $X$ to the corresponding styled image $X_S$.
The style template $S$ is added to the input image.
\adv aims to steal the styling functionality of \victimfnc that maps the input image to the corresponding styled image without explicit access to the secret styling template $S$.
We summarise the notation in~\Cref{tab:notations}.

\setlength\tabcolsep{4pt}
\begin{table}
\caption{Summary of the notation used throughout this work.}
\centering
\begin{tabular}{llll}
    \hline
    \victim         & victim                             & $X$                 & unstyled image(s)\\
    \adv            & adversary                          & $X_S$               & styled image(s)\\
    \victimfnc      & victim model                       & $X_{\mathcal{A}}$   & \adv's unstyled image(s)\\
    \advfnc         & surrogate model                    & $X_{\mathcal{A}_S}$ & \adv's styled images\\
    $S$             & style template                     & $X_{lr}/X_{hr}$     & low/high resolution image(s)\\
    $S_X$           & source style images                & $X_{\mathcal{A}_{lr}}/X_{\mathcal{A}_{hr}}$ & \adv's low/high resolution image(s)\\
    \hline
\end{tabular}\label{tab:notations}
\end{table}

\subsection{Attack Methodology}
\adv queries \victimfnc using \victim's API to obtain the corresponding styled $X_{\mathcal{A}_S}$ images generated by \victimfnc.
This in consistent with the practical settings likely to be encountered in real world applications.
This pair of unstyled input and styled output image $(X_{\mathcal{A}}, X_{\mathcal{A}_S})$ constitutes the surrogate training data used by \adv to train the surrogate model \advfnc.
Since \adv relies on training \advfnc using pairs of data points, we use paired image translation models as our attack model.
As previously explained, training paired translation models is not realistic in many real world applications due to the lack of styled images.
However in our case, \adv leverages the fact that they can obtain many $X_{\mathcal{A}_S}$ by querying $F_\mathcal{V}$.
Additionally, our attack makes no assumptions regarding the styling technique used to generate the styled images. However, models that we target do require GANs for high quality transformations.
Even though the styling template $S$ applied to unstyled input images by \victimfnc is learned explicitly from images, it can be stolen and implicitly embedded in \advfnc.

More formally, from the set of possible mappings $\mathcal{F}$, the mapping $F_\mathcal{V} : X \rightarrow X_S$ is optimal if according to some quantitative criterion $\mathcal{M}_S$, the resulting transformation captures the style $S$ by maximizing $\mathcal{M}_S(X_S, S)$:
{\small
\begin{equation}
    F_\mathcal{V} = arg\underset{F'_\mathcal{V} \in \mathcal{F}}{\max} \mathcal{M}_S(X_S, S)
\end{equation}}

as well as retains the semantics of the input image $\mathcal{M}_X(X, X_S)$ by maximizing another criterion $\mathcal{M}_X$:
{\small
\begin{equation}
    F_\mathcal{V} = arg\underset{F'_\mathcal{V} \in \mathcal{F}}{\max} \mathcal{M}_X(X, X_S)
\end{equation}}

\adv's goal is to learn the same kind of mapping $F_\mathcal{A}: X_\mathcal{A} \rightarrow X_{\mathcal{A}_S}$ using images obtained from \victimfnc.
This transformation must be successful according to both criteria $\mathcal{M}_X(X_\mathcal{A}, X_{\mathcal{A}_S})$ and $\mathcal{M}_S(X_{\mathcal{A}_S}, S)$:

{\small
\begin{align}
    \nonumber F_\mathcal{A} &= arg\underset{F'_\mathcal{A} \in \mathcal{F}}{\max} \mathcal{M}_X(X_\mathcal{A}, X_{\mathcal{A}_S})\quad and\\
    F_\mathcal{A} &= arg\underset{F'_\mathcal{A} \in \mathcal{F}}{\max} \mathcal{M}_S(X_{\mathcal{A}_S}, S)
\end{align}}

However, \adv does not have access to $S_X$, and thus the template $S$.
Instead, they use a proxy metric $\mathcal{M}_\mathcal{P}(F_\mathcal{V}, F_\mathcal{A}, X)$.
Abusing notation, we define it as:

{\small
\begin{align}\label{proxy_metric}
    \nonumber &F_\mathcal{A} = arg\underset{F'_\mathcal{A} \in \mathcal{F}}{\max} M_\mathcal{P}(F_\mathcal{V}, F_\mathcal{A}, X) \quad s.t.\\
    \nonumber &if\quad F_\mathcal{A}(X) \approx F_\mathcal{V}(X) \quad then\\
    \nonumber &F_\mathcal{A} = arg\underset{F'_\mathcal{A} \in \mathcal{F}}{\max} \mathcal{M}_X(X_\mathcal{A}, X_{\mathcal{A}_S})\quad and\\
    &F_\mathcal{A} = arg\underset{F'_\mathcal{A} \in \mathcal{F}}{\max} \mathcal{M}_S(X_{\mathcal{A}_S}, S)
\end{align}}

In other words, we assume that if images produced by \advfnc are similar to those produced by \victimfnc, then \advfnc has successfully learned the style $S$.

We apply our attacks to the following three settings.

\noindent\textbf{Setting 1: \mtop.}
\mtop is a style transfer task in which a picture of a scene is transformed to look as if it was painted in Claude Monet's signature style.
We train \victimfnc as an unpaired image to image translation problem, i.e, the training data consists of unordered and uncorrelated photos of landscapes, and Monet's paintings $S_X$ that contains his signature style $S$.
$S_X$ is secret and not visible to \adv.
The model learns the features in the Monet paintings and applies them to translate generic photos.
\adv queries \victimfnc with images $X_{\mathcal{A}}$ from a different dataset than \victim's training data and obtains corresponding $X_{\mathcal{A}_S} =  F_\mathcal{V}(X_{\mathcal{A}})$.
\advfnc is trained using this generated dataset $(X_{\mathcal{A}}, X_{\mathcal{A}_S})$ as a paired image to image translation problem between the two domains.

\noindent\textbf{Setting 2: \stoa.}
\stoa is used as part of a real world web application which allows users to upload their selfie and convert it to a corresponding anime image~\citep{selfie2animeService}.
We do not attack the actual web interface due to its terms of service on reverse engineering and scraping.
However, the authors disclose the data and architecture~\citep{ugatit} used to train their model, and we use it to train our \victimfnc.
The training data is unpaired - it consists of a training set of selfie images, and a styling template $S$ consisting of anime images $S_X$.
\adv queries \victimfnc with images $X_{\mathcal{A}}$ from a different dataset than \victim's training data and obtains corresponding $X_{\mathcal{A}_S} =  F_\mathcal{V}(X_{\mathcal{A}})$.
\advfnc is trained using this generated dataset $(X_{\mathcal{A}}, X_{\mathcal{A}_S})$ as a paired image to image translation problem between the two domains.

\noindent\textbf{Setting 3: \superres.}
In \superres a low resolution image is upscaled to its high resolution counterpart using a GAN.
We consider mapping low resolution images, (using a 4x bicubic downscaling) to their high resolution counterparts with 4x upscaling.
\victimfnc's generator is trained in a supervised fashion using a paired dataset $(X_{lr}, X_{hr})$ of low resolution and corresponding high resolution images. Here $X_{hr}$ corresponds to $X_S$.
There is no explicit styling template $S$.
\adv queries \victimfnc with publicly available low resolution images $X_{\mathcal{A}_{lr}}$ to obtain the corresponding high resolution image $X_{\mathcal{A}_{hr}} = F_\mathcal{V}(X_{\mathcal{A}_{lr}})$.
This paired surrogate data $(X_{\mathcal{A}_{lr}}, X_{\mathcal{A}_{hr}})$ is used to train \advfnc.

Our attack does not make any assumptions about the training or the architecture of \victimfnc, and it relies solely on the query access.
Hence, it is applicable to other image translation settings and can be used to extract \victimfnc{s} with a well-defined style $S$.

\section{Evaluation}\label{evaluation}

\subsection{Datasets}\label{datasets}

\noindent\textbf{\mtop.}
\victimfnc is trained using the Monet2Photo~\citep{Monet2Photo} dataset consisting of 1193 Monet paintings and 7038 natural photos divided into train and test sets.
For training \advfnc, we use the Intel Image Classification dataset~\citep{intelic}.
It consists of 14,034 of images of various nature and city landscapes.
As the benchmark dataset, we choose the Landscape~\citep{landscape} dataset that contains 13,233 images from the same domain but was not used to train either model.

\noindent\textbf{\stoa.}
\victimfnc is trained on the selfie2anime dataset~\citep{selfie2anime} which comprises 3400 selfies and 3400 anime faces as part of the training data, and 100 selfies and 100 anime faces in the test set.
\adv uses the CelebA~\citep{liu2015faceattributes} dataset for attacking \victim which consists of 162,770 photos of celebrities.
For the benchmark dataset, we use the LFW dataset~\citep{LFWTech} which contains a test set of 13233 selfie images.

\noindent\textbf{\superres.}
\victimfnc is trained using the DIV2K~\citep{Agustsson_2017_CVPR_Workshops} dataset that consists of 800 pairs of low and corresponding high resolution images in the training set, and 200 image pairs for validation and testing.
\adv uses the FLICKR2K dataset for the extraction.
It consists of 2650 train (and 294 test images).
We create the SRBechmark dataset by combining multiple benchmarking datasets used for super resolution model evaluations including SET5~\citep{LapSRN}, SET14~\citep{LapSRN}, URBAN100~\citep{LapSRN} and BSD100~\citep{LapSRN} containing 294 images. All high resolution images are 4x bicubic downsampled.

\setlength\tabcolsep{4pt}
\begin{table}
    \centering
    \caption{Datasets used to for training \victimfnc and \advfnc.}
    \label{tab:data}
    \small
    \begin{tabular}{c | c c | c | c} \hline
     Task            & \multicolumn{2}{c|}{\victim Data} & \adv Data &  Benchmark Data\\ \hline
     \mtop  & \multicolumn{2}{c|}{Monet2Photo}  & Intel     & Landscape\\
                     & Train & Test                      & Train     & Test\\
                     & 6287  & 751                       & 14034     & 4319\\ \hline
    \stoa  & \multicolumn{2}{c|}{Selfie2Anime} & CelebA    & LFW\\
                     & Train & Test                      & Train     & Test\\
                     & 3400  & 100                       & 162770    & 13233 \\ \hline
    \superres & \multicolumn{2}{c|}{DIV2K}        & FLICKR2K  & SRBenchmark\\
                     & Train & Test                      & Train     & Test\\
                     & 800 & 200                         & 2650      & 294\\
     \hline
    \end{tabular}
\end{table}

\subsection{Architectures}\label{architectures}

For \textbf{\mtop},
we use the state of the art CycleGAN~\citep{cyclegan} architecture as the \victimfnc architecture.
For \advfnc architecture we use a paired image-to-image translation model Pix2Pix~\citep{pix2pix}.
For \textbf{\stoa},
we use the UGATIT generative model~\citep{ugatit} as the \victimfnc architecture which is similar to a CycleGAN.
This is the model used by the Selfie2Anime service~\citep{selfie2animeService}.
Like for \mtop, we use a paired image-to-image translation model Pix2Pix~\citep{pix2pix} for \advfnc.
For \textbf{\superres},
we use the state of the art SRGAN model~\citep{srgan} or the \victimfnc architecture, and for \advfnc we use a similar SRResNet model~\citep{srgan} with ResNet convolutional units.

\subsection{Experiments}\label{experiments}

\Cref{tab:models-fid-monet,tab:models-fid-selfie,tab:models-psnr-ssim-superres} summarize the results of our experiments.
We report the mean value for a given distance metric, and standard deviations where applicable.
In Tables~\ref{tab:models-fid-monet} and \ref{tab:models-fid-selfie} we report the effectiveness of the attack measured using the FID score for style transfer settings.
In Table~\ref{tab:models-psnr-ssim-superres} we report the effectiveness of the attack measured using PSNR and SSIM for the \superres setting.
In each setting for each dataset, we report on three different experiments:
\begin{enumerate}[label=\textbf{(\Alph*)}]
    \item \label{exp1} \textbf{effectiveness of \victimfnc transformations:} this is the baseline performance of \victimfnc with respect to the original test set (cell~\textbf{\ref{exp1}} in~\Cref{tab:models-fid-monet,tab:models-fid-selfie,tab:models-psnr-ssim-superres}).
    \item \label{exp2} \textbf{comparison of \advfnc and \victimfnc:} this is measured using the proxy metric (Equation~\ref{proxy_metric}) that \adv uses to assess the effectiveness of \advfnc in terms of how closely its outputs resemble the outputs of \victimfnc (cell~\textbf{\ref{exp2}} in~\Cref{tab:models-fid-monet,tab:models-fid-selfie,tab:models-psnr-ssim-superres}).
    \item  \label{exp3} \textbf{effectiveness of \advfnc transformations:} this is measured the same way as~\textbf{\ref{exp1}}. Note that \adv cannot directly compute this effectiveness metric because they do not have access to the ground truth (in the form of the styling set $S$ or paired images in the case of \superres; cell~\textbf{\ref{exp3}} in~\Cref{tab:models-fid-monet,tab:models-fid-selfie,tab:models-psnr-ssim-superres}).
\end{enumerate}

For \mtop, we can see that our attack is effective at extracting \victimfnc.
In Table~\ref{tab:models-fid-monet}, we report the FID scores for the three experiments using two test sets.
In all cases, we see that \advfnc performs comparably to \victimfnc.

\begin{figure}[t]
    \resizebox{1.\columnwidth}!{
    \begin{tabular}{ccc}
        \includegraphics[width=0.25\columnwidth]{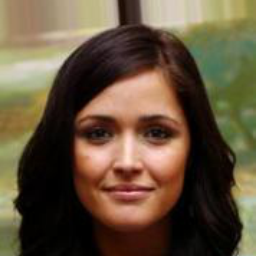} &
        \includegraphics[width=0.25\columnwidth]{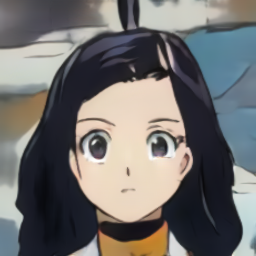} &
        \includegraphics[width=0.25\columnwidth]{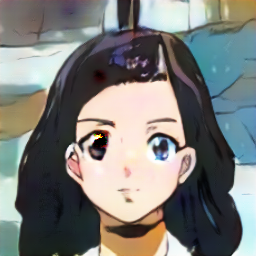} \\
    \end{tabular}}
    \caption{Example of successful transformations produced by \victimfnc and \advfnc that are visually similar.}
    \label{fig:example_same}
\end{figure}

\begin{figure}[t]
    \resizebox{1.\columnwidth}!{
    \begin{tabular}{ccc}
        \includegraphics[width=0.25\columnwidth]{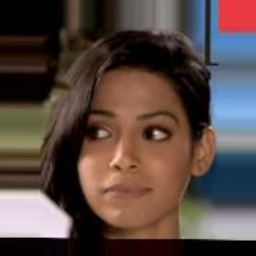} &
        \includegraphics[width=0.25\columnwidth]{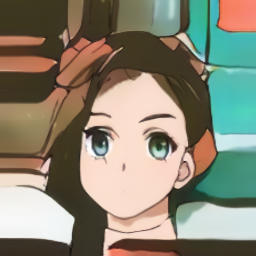} &
        \includegraphics[width=0.25\columnwidth]{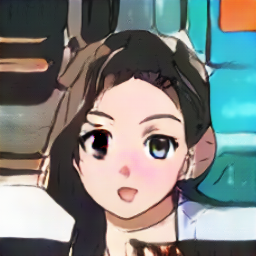} \\
    \end{tabular}}
    \caption{\victimfnc and \advfnc learn different representations and can produce transformations that have different features even though both transformations are comparably effective.}
    \label{fig:example_diff}
\end{figure}

Our attack is effective at extracting a \stoa \victimfnc.
In Table~\ref{tab:models-fid-selfie}, we report the FID scores for the three experiments, using two test sets.
We observe a large FID score for the \victim's test set (Selfie2Anime).
However, using a benchmark test set (LFW), we observe that both models compare similarly and produce similar styled images, c.f. Figure~\ref{fig:example_same}.
Additionally, we observe that many transformations are qualitatively similar (both models successfully produce anime faces) even though the resulting transformations are different, c.f. Figure~\ref{fig:example_diff}.

Finally, our attack is effective at extracting the \superres \victimfnc, c.f. Table~\ref{tab:models-psnr-ssim-superres}.
Both in the case of SSIM (Table~\ref{tab:models-ssim-superres}) and PSNR (Table~\ref{tab:models-psnr-superres}), \advfnc performs similarly to \victimfnc.
This holds for the proxy comparison between \victimfnc and \advfnc, as well as w.r.t. the original high resolution images.

\setlength\tabcolsep{4pt}
\begin{table}[t]
    \centering
    \caption{FID of \victimfnc and \advfnc for \mtop. We report the mean and standard deviation over five runs.}
    \label{tab:models-fid-monet}
        \small
    \begin{tabular}{c|c|c c|c c} \hline
                                      &                 & \multicolumn{4}{c}{Compared Distribution}\\
     Output Distribution              & Test Set        & & \victim Output  & & Monet Paintings\\ \hline
     \multirow{2}{*}{\victim Output}  & Monet2Photo Test & & N/A             & \multirow{2}{*}{~\textbf{\ref{exp1}}} & $58.05\pm0.74$\\
                                      & Landscape       & & N/A             & & $49.59\pm0.82$\\\hline
     \multirow{2}{*}{\adv Output}     & Monet2Photo Test & \multirow{2}{*}{~\textbf{\ref{exp2}}} & $42.86\pm1.01$  & \multirow{2}{*}{~\textbf{\ref{exp3}}} & $61.62\pm1.05$\\
                                      & Landscape       & & $15.38\pm0.54$           & & $53.99\pm1.21$\\\hline
    \end{tabular}
\end{table}

\setlength\tabcolsep{4pt}
\begin{table}[t]
    \centering
    \caption{FID of \victimfnc and \advfnc for \stoa. We report the mean and standard deviation over five runs.}
    \label{tab:models-fid-selfie}
    \small
    \begin{tabular}{c|c|c c|c c} \hline
                                      &                   & \multicolumn{4}{c}{Compared Distribution}\\
     Output Distribution              & Test Set     & & \victim Output  & & Anime Faces\\ \hline
     \multirow{2}{*}{\victim Output}  & Selfie2Anime & & N/A             & \multirow{2}{*}{~\textbf{\ref{exp1}}} & $69.02\pm1.91$\\
                                      & LFW Test     & & N/A             & & $56.06\pm2.47$\\\hline 
     \multirow{2}{*}{\adv Output}     & Selfie2Anime & \multirow{2}{*}{~\textbf{\ref{exp2}}} & $106.63\pm5.53$ & \multirow{2}{*}{~\textbf{\ref{exp3}}} & $108.68\pm3.19$\\
                                      & LFW Test     &  & $19.67\pm0.69$           & & $69.42\pm2.04$\\\hline 
    \end{tabular}
\end{table}

\setlength\tabcolsep{4pt}
\begin{table}[t]
    \caption{PSRN and SSIM of \victimfnc and \advfnc for \superres. We report the mean and standard deviation over five runs.}
    \label{tab:models-psnr-ssim-superres}
    \begin{subtable}{\linewidth}\centering
    \caption{\textbf{SSIM}. Comparing \advfnc to \victimfnc captures the true performance.}\label{tab:models-ssim-superres}
        \small
    \begin{tabular}{c|c|c c|c c} \hline
                                      &                   & \multicolumn{4}{c}{Compared Distribution}\\
     Output Distribution              & Test Set          & & \victim Output  & & Original high-res\\ \hline
     \multirow{2}{*}{\victim Output}  & DIV2K Test        & & N/A             & \multirow{2}{*}{~\textbf{\ref{exp1}}} & $0.74 \pm 0.05$\\
                                      & SRBENCHMARK  & & N/A             & & $0.69 \pm 0.12$\\\hline
     \multirow{2}{*}{\adv Output}     & DIV2K Test        & \multirow{2}{*}{~\textbf{\ref{exp2}}} & $0.75 \pm 0.08$ &  \multirow{2}{*}{~\textbf{\ref{exp3}}} & $0.80 \pm 0.10$\\
                                      & SRBENCHMARK  &  & $0.76 \pm 0.09$ &  & $0.61 \pm 0.06$\\\hline
    \end{tabular}
    \end{subtable}\\
    \begin{subtable}{\linewidth}\centering
    \caption{\textbf{PSNR}. Comparing \advfnc to \victimfnc captures the true performance.}\label{tab:models-psnr-superres}
        \small
    \begin{tabular}{c|c|c c|c c} \hline
                                      &                   & \multicolumn{4}{c}{Compared Distribution}\\
     Output Distribution              & Test Set          & & \victim Output   & & Original high-res\\ \hline
     \multirow{2}{*}{\victim Output}  & DIV2K Test        & & N/A              & \multirow{2}{*}{~\textbf{\ref{exp1}}} & $24.67 \pm 1.81$\\
                                      & SRBENCHMARK  & & N/A              & & $22.59 \pm 4.51$\\\hline
     \multirow{2}{*}{\adv Output}     & DIV2K Test        & \multirow{2}{*}{~\textbf{\ref{exp2}}} & $24.74 \pm 5.04$ & \multirow{2}{*}{~\textbf{\ref{exp3}}} & $23.24 \pm 3.77$\\
                                      & SRBENCHMARK  & & $20.64 \pm 3.33$ & &  $18.13 \pm 3.61$\\\hline
    \end{tabular}
    \end{subtable}
\end{table}

\subsection{User Study}\label{userstudy}

Since distance metrics do not necessarily capture human perception, we conducted an extensive user study to evaluate how humans rate the image translation tasks.
All user study materials are available in the supplement.

\noindent\textbf{Materials.}
We designed a questionnaire in Google Forms which we shared with the participants.

\begin{figure*}[t]
    \resizebox{1.\columnwidth}!{
    \begin{tabular}{lcr}
        \includegraphics[width=0.66\columnwidth]{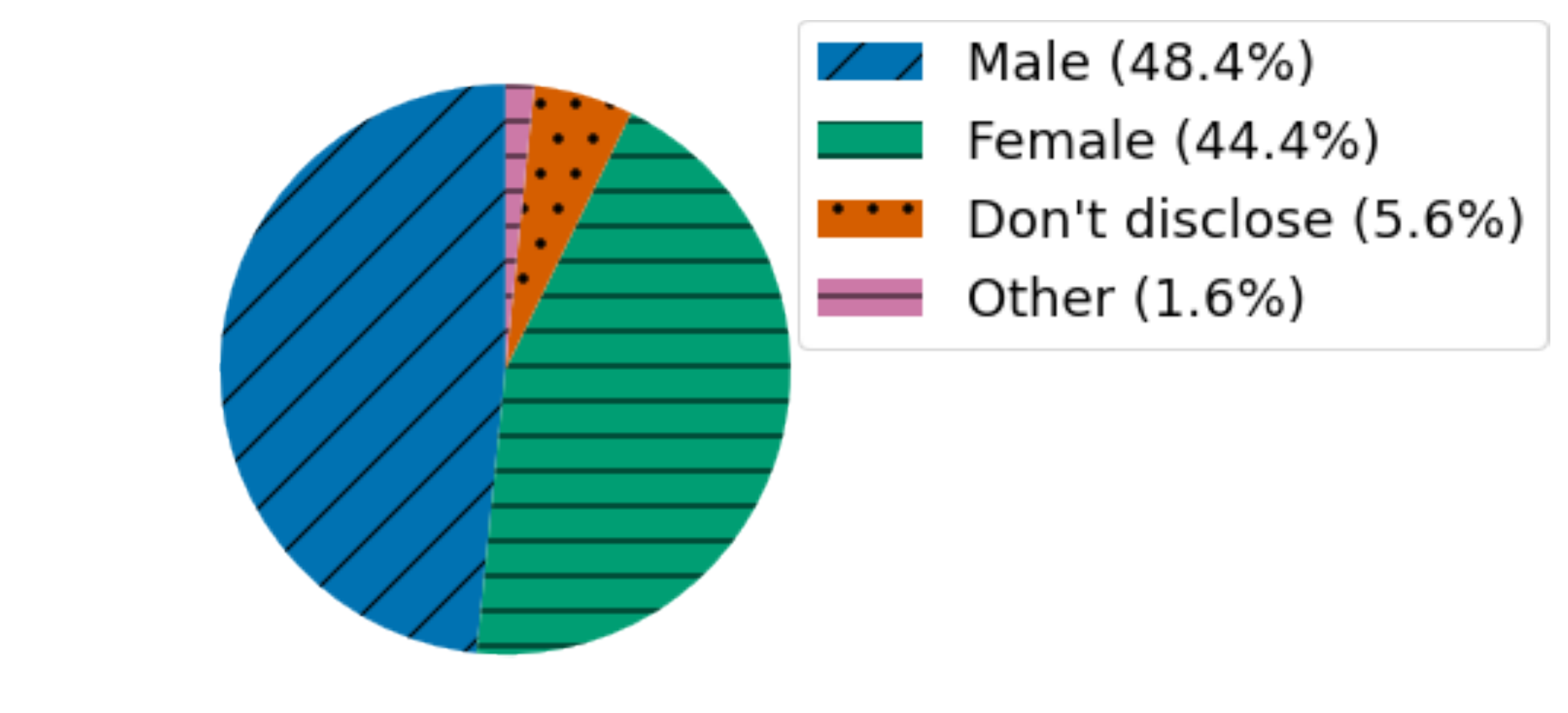} &
        \includegraphics[width=0.66\columnwidth]{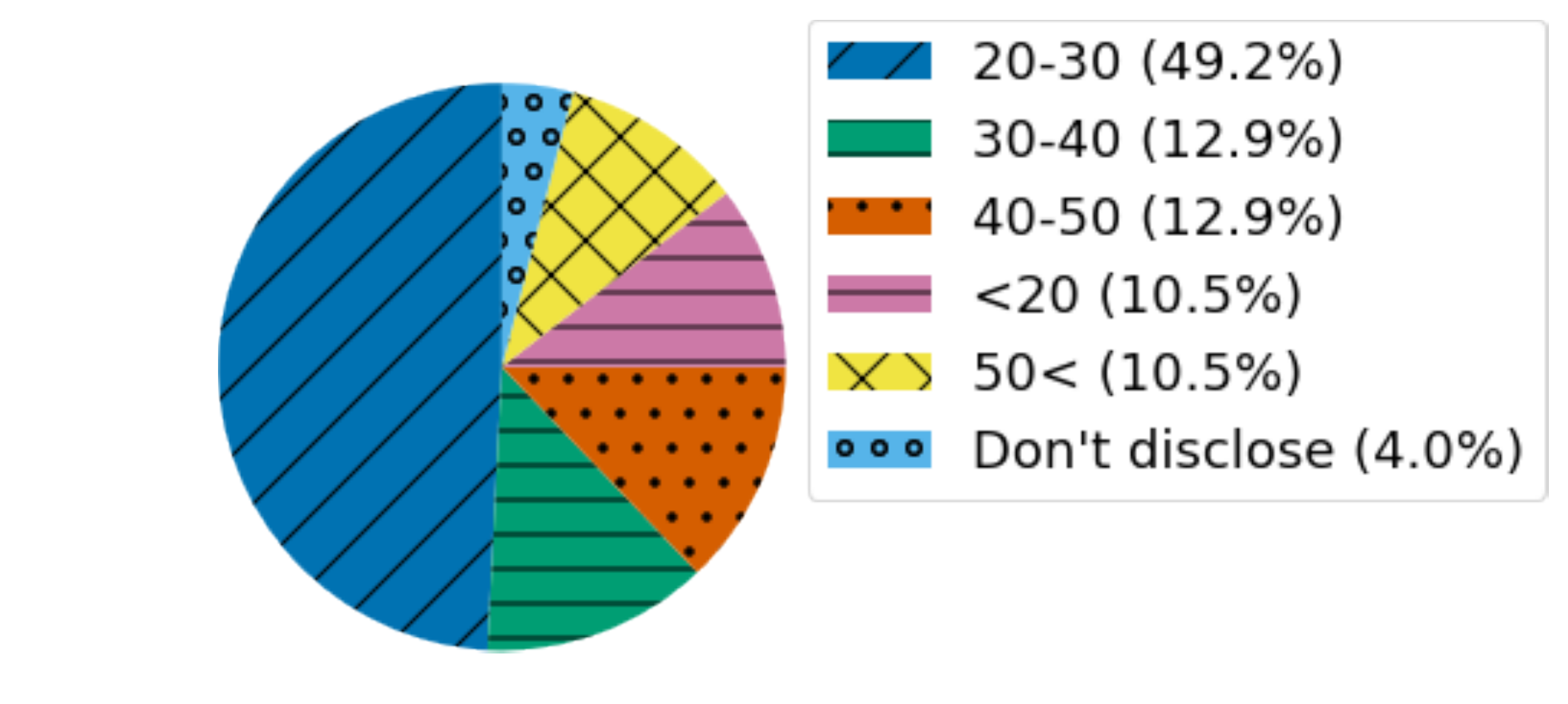} &
        \includegraphics[width=0.66\columnwidth]{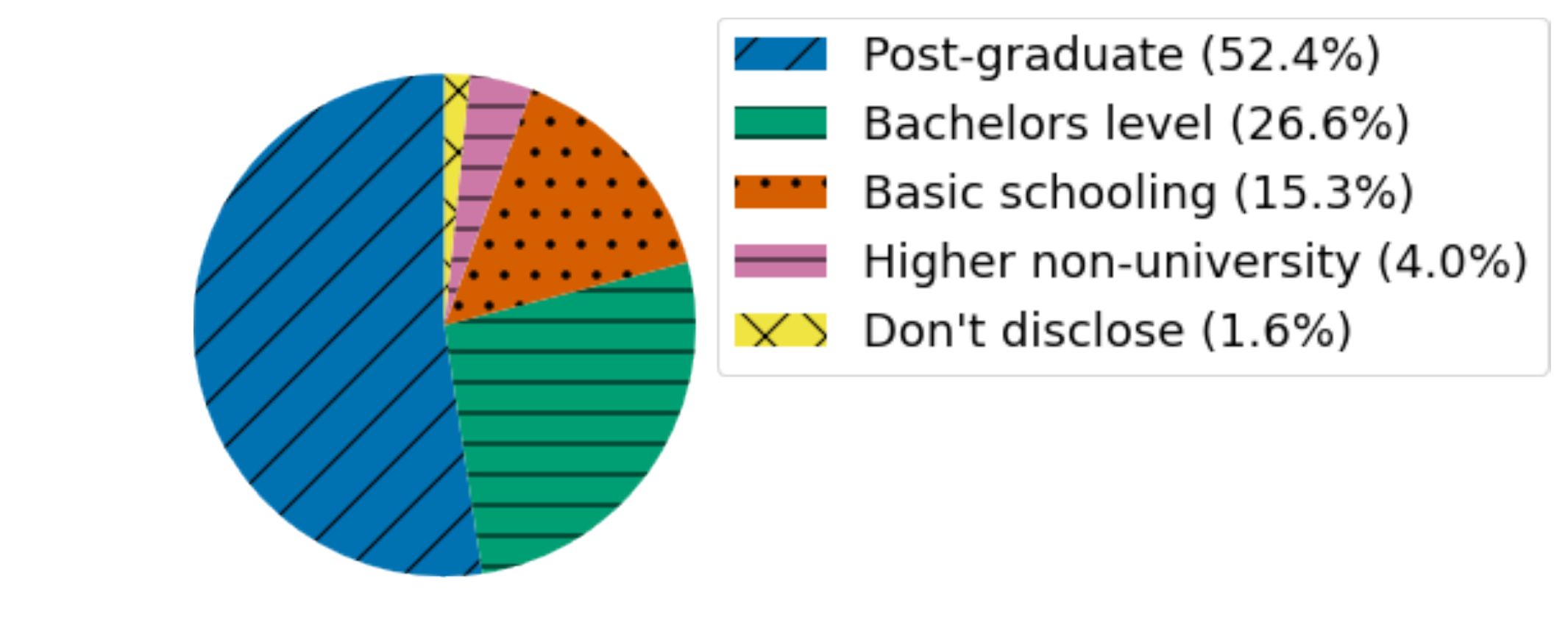} \\
    \end{tabular}}
    \caption{Distribution of participant demographics. Left to right: gender, age and highest-completed education.}
    \label{fig:demographics}
\end{figure*}

\noindent\textbf{Participants.}
We conducted the user study with 125 participants
invited through social media (Twitter, LinkedIn, Facebook) and institutional channels (Slack, Teams, Telegram groups).
We collected basic demographics about the participants, including age, highest obtained education, and gender.
Our user study data collection conforms to our institution's data management guidelines.
No personally identifiable information was collected as part of the study.
\Cref{fig:demographics} represents the demographics of the participants of the study.

\noindent\textbf{Design of User Study.}
In the questionnaire, we first introduced the problem of image style transfer with examples of Monet paintings and anime images.
We explained that the study aims to evaluate which of the shown style transformations are successful in the participant's opinion, and does not evaluate their skill in making this judgement.
Participants then evaluated the style transformations using a Likert scale from 1 to 5.
A score of 5 indicated that the result looked very similar to a Monet painting of the input scene or an anime character corresponding to the source selfie photo, and a score of 1 that there was no such resemblance.

\noindent\textbf{Procedure.}
We conducted the user study on two style transfer settings.
For both tasks, we included 20 pairs consisting of an original unstyled image and its corresponding styled variant.
Half of the them (10) were styled with \victimfnc and the other half with \advfnc.
The pairs were shown in random order.

\begin{figure}[t]
    \resizebox{1.\columnwidth}!{
    \begin{tabular}{lccr}
        \includegraphics[width=0.5\columnwidth]{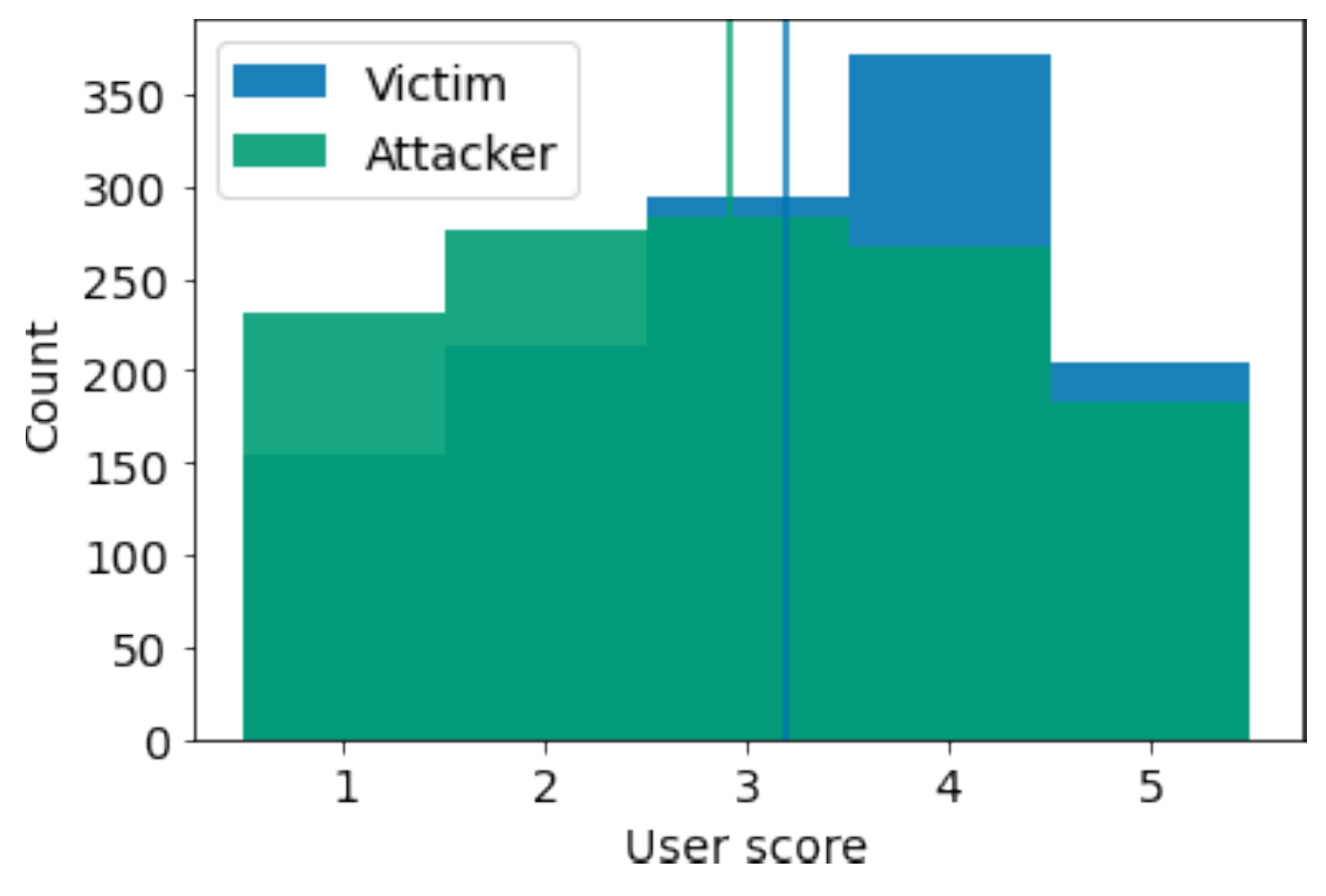} &
        \includegraphics[width=0.5\columnwidth]{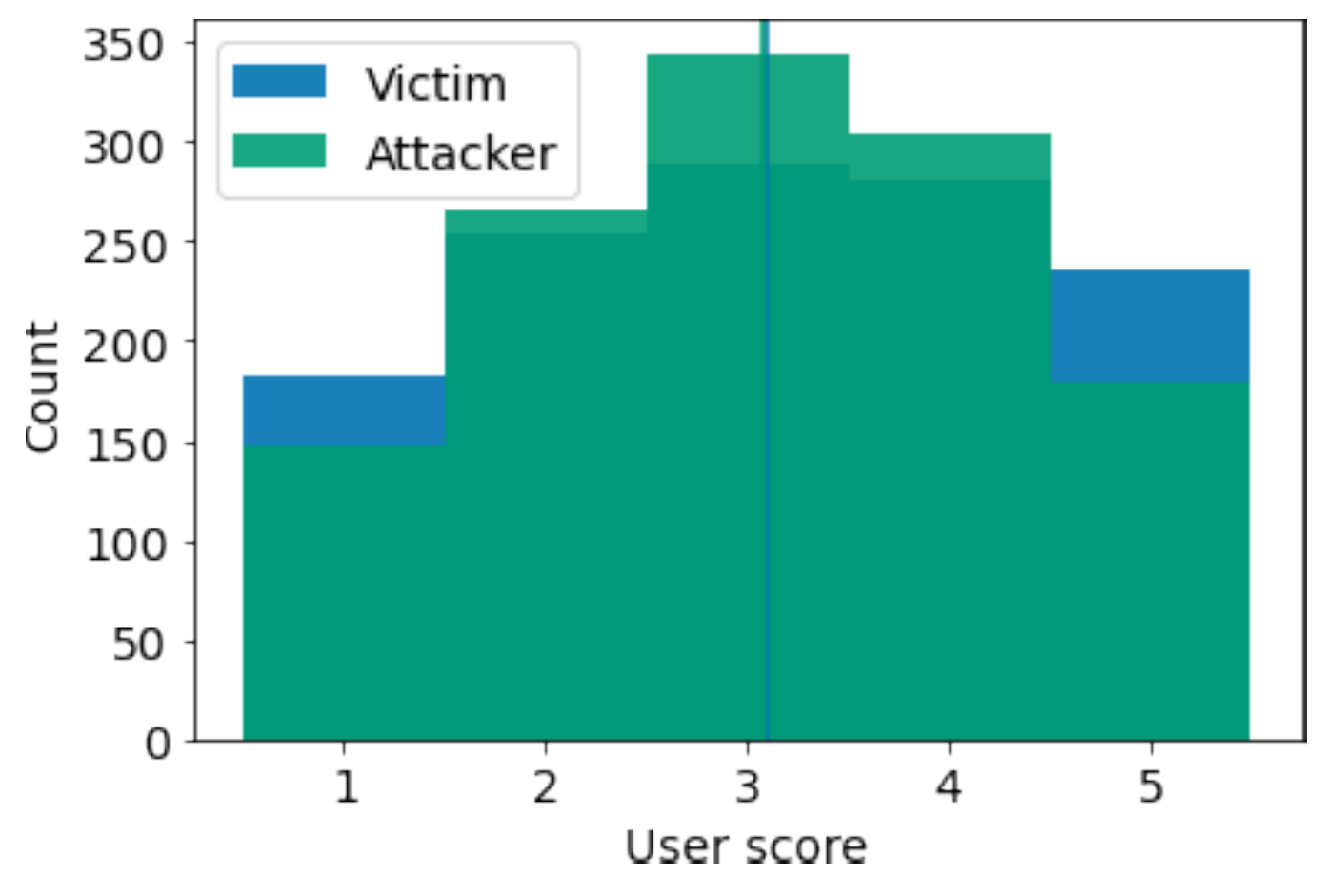}
    \end{tabular}}
    \caption{Comparison of scores assigned by the user study participants. \textbf{\mtop (left)}: for \victimfnc score mean~$=3.20$ and std~$=1.59$, and for \advfnc mean~$=2.91$ and std~$=1.76$. Vertical lines indicate means of corresponding distributions. \textbf{\stoa (right)}: for \victimfnc score mean~$=3.11$ and std~$=1.76$, and for \advfnc mean~$=3.08$ and std~$=1.50$. Vertical lines indicate means of corresponding distributions.}
    \label{fig:study}
\end{figure}

\noindent\textbf{Results.}
\Cref{fig:study} shows the distribution of scores for \victimfnc and \advfnc in the \mtop and \stoa test settings, respectively.
Our evaluation concerns whether the differences between results obtained from \victimfnc and \advfnc were \emph{statistically significant}.
We first conducted a \emph{t-test}, which yields a conditional probability (\emph{$p$-value}) for receiving values deviating equally or more between the sample statistics as observed in the experiments, under the \emph{null hypothesis} of equal underlying population distributions.
The outcome is considered statistically significant if $p$ exceeds the \emph{significance threshold} $\alpha$, which we set to $0.05$.
Since sample variances were unequal between \victimfnc and \advfnc, we used Welch's t-test instead of Student's t-test.

The null hypothesis of equivalent population distributions was rejected for \mtop (\victimfnc's score mean~$=3.20$ and std~$=1.59$; \advfnc's score mean~$=2.91$ and std~$=1.76$; $p=1.8e-8$)
but not for \stoa (\victimfnc's score mean~$=3.11$ and std~$=1.76$; \advfnc's score mean~$=3.08$ and std~$=1.50$; $p=0.63$).
In other words, the probability of obtaining results deviating between \victimfnc and \advfnc as much as observed (or more) would be very high for \stoa but very low for \mtop, if \victimfnc and \advfnc indeed performed similarly.

However, the t-test alone is insufficient to evaluate the divergence between \victimfnc and \advfnc.
We additionally performed a \emph{two one-sided t-test} (TOST) for \emph{equivalence} to test whether the results fall between \emph{equivalence bounds}.
As the bounds, we chose the range $[-0.3, 0.3]$ for \emph{Cohen's $d$}, which is the mean difference between the models' scores standardized by their pooled standard deviations~\citep{cohen1988}.
The null hypothesis is reversed from the standard t-test, now assuming the \emph{non-equivalence} between underlying population distributions.
The null hypothesis is rejected if the observed difference falls within the equivalence bounds.

For \stoa the null hypothesis was rejected within the equivalence bounds of $[-0.11, 0.11]$ ($p=0.0476$),
and for \mtop within $[-0.38, 0.38]$ ($p=0.04829$), using $\alpha=0.05$.
Both fall within $[-0.3, 0.3]$ for Cohen's $d$, which corresponds to the raw value range of approximately $[-0.50, 0.50]$ in \stoa and $[-0.49, 0.49]$ in \mtop.
Hence, we reject the non-equivalence of \victimfnc and \advfnc in \textbf{both} test settings.

While a statistically significant difference between \victimfnc and \advfnc was observed in \mtop, it was too small to count as sufficiently large for meaningful non-equivalence in TOST.
In \stoa, both the t-test and TOST supported rejecting the non-equivalence of \victimfnc and \advfnc.
We therefore conclude that \textbf{\advfnc successfully achieved performance close to \victimfnc based on human judgement}.



\section{Attack Efficiency}\label{efficiency}

Having established the baseline performance of the attack that uses the entire dataset, we now focus on its efficiency.
In~\Cref{variable-data} we study how reducing the amount of data affects the performance of \advfnc.
In~\Cref{data-augmentation} we explore different data augmentation techniques to further reduce the number of queries required.

\subsection{Variable Amount of Data}\label{variable-data}

\begin{figure}[t]
    \resizebox{1.\columnwidth}!{
    \begin{tabular}{lcr}
        \includegraphics[width=0.4\columnwidth]{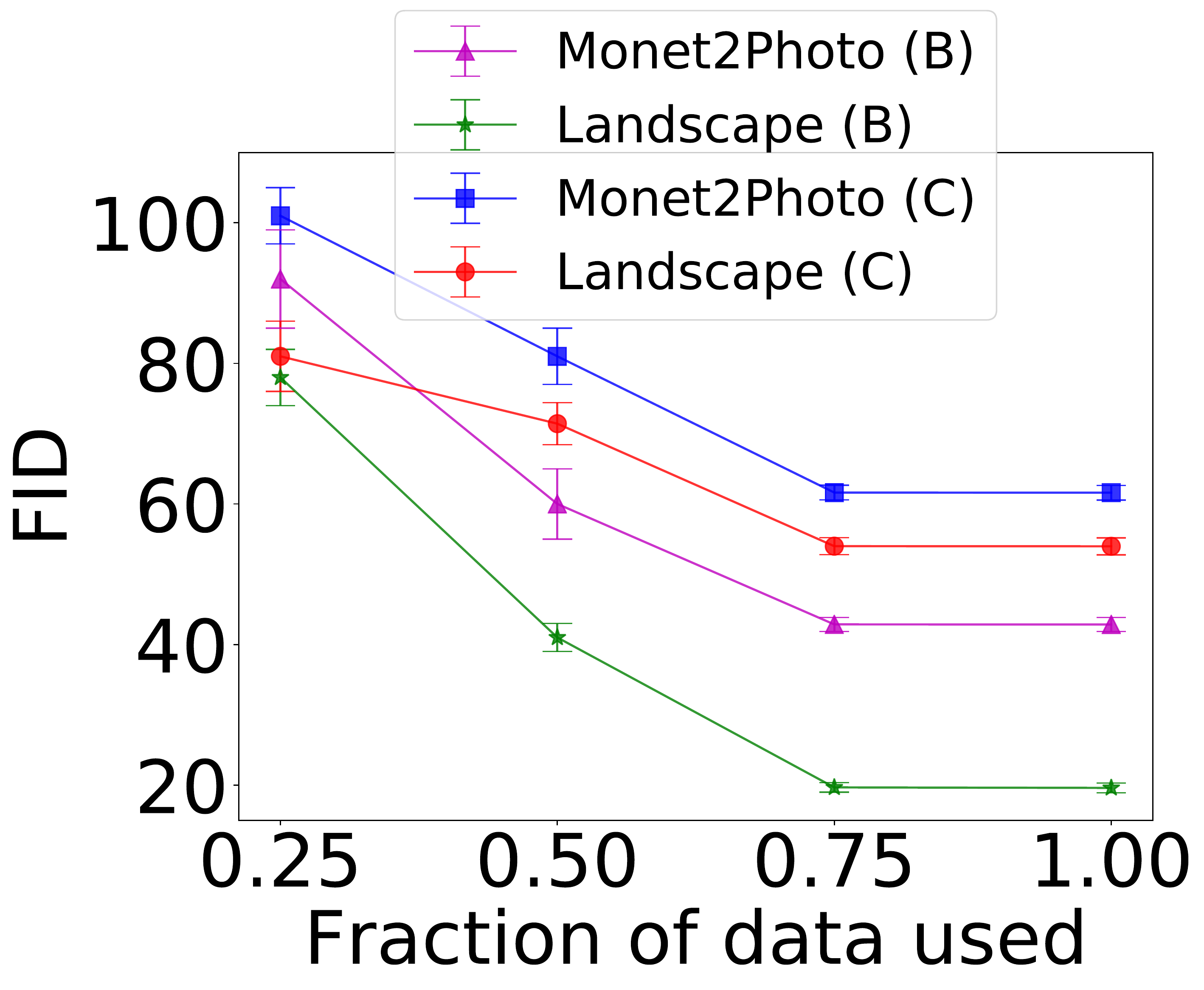} &
        \includegraphics[width=0.4\columnwidth]{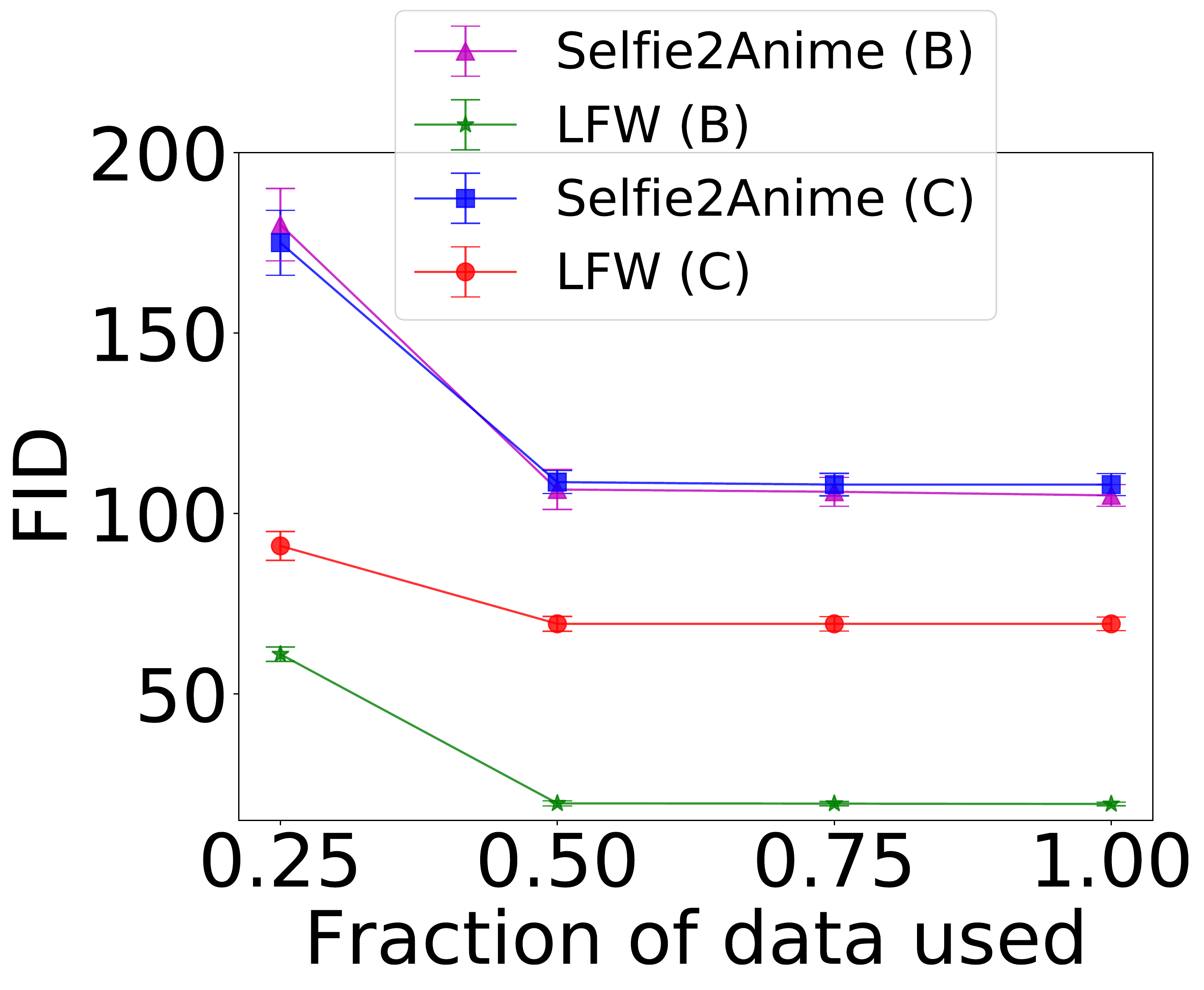} &
        \includegraphics[width=0.4\columnwidth]{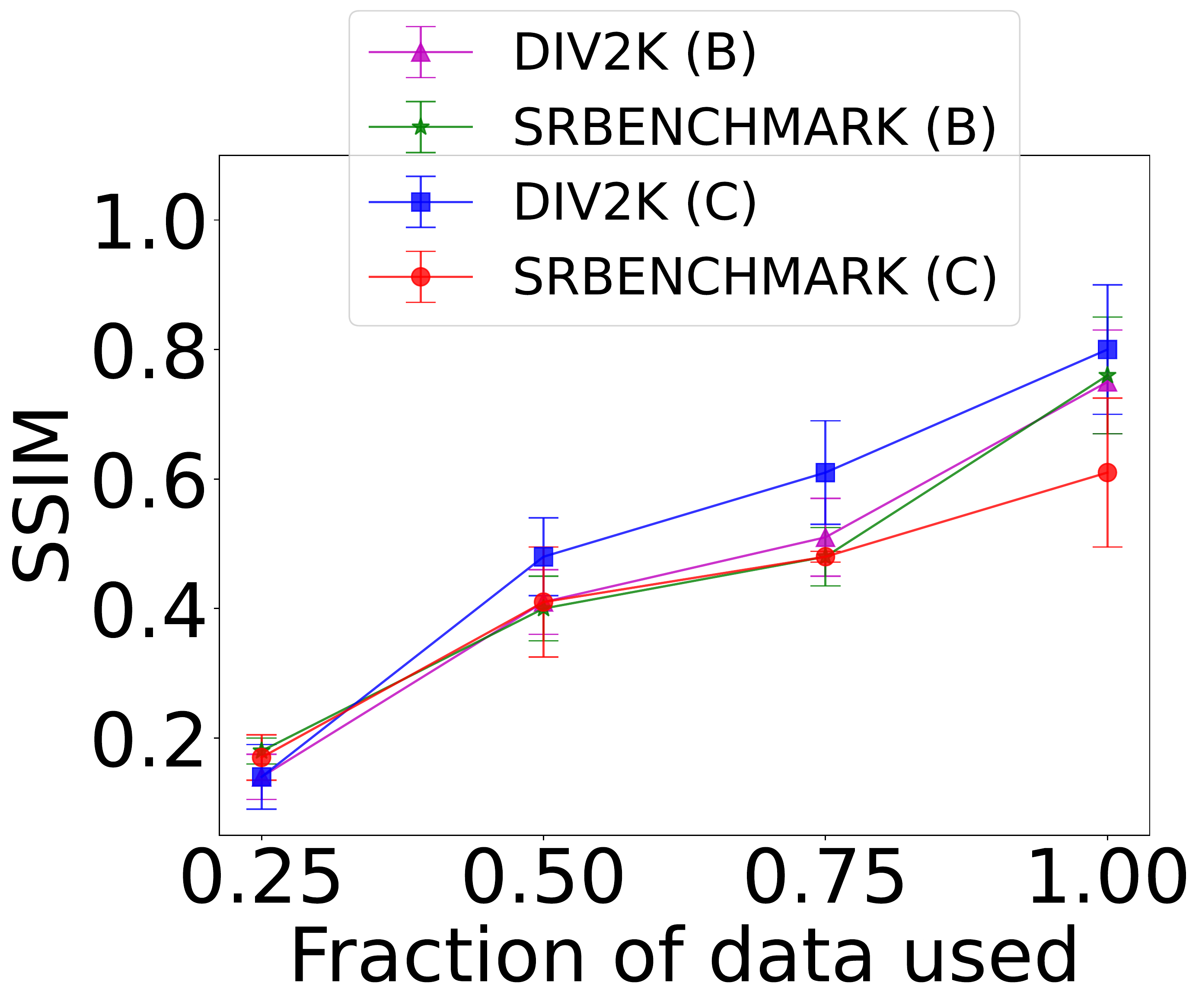}
    \end{tabular}}
    \caption{Impact of the amount of data on the effectiveness of the attack. Left to right: \mtop, \stoa, \superres (SSIM). For \mtop the performance plateaus at $75\%$, \stoa stops improving at $50\%$, \superres continues to improve up to $100\%$ (we observe the same trend for SSIM and PSNR).}
    \label{fig:variable-data}
\end{figure}

For each task, we reduce the amount of data used to extract \victimfnc.
We conduct the experiment with randomly sampled $25\%$, $50\%$, $75\%$ of the original attack dataset; each experiment is repeated five times.
\Cref{fig:variable-data} depicts how the performance of \advfnc decreases with less data.
We plot both the comparison with \victimfnc's output (experiment~\ref{exp2}) and with the style images (experiment~\ref{exp3}).

For \mtop, the model does not improve with more than $75\%$.
We conjecture that this is enough because the style target style (Monet paintings) is consistent across many images.

For \stoa, the improvement plateaus around $50\%$.
Although the mean FID does not improve with more data, the standard deviation is lower at $75\%$.
It is the most difficult of the three tasks since there is substantial variability in the anime faces.
However, the attack dataset is the largest and it has a lot of redundancy.

For \superres, the attack improves with more data, up to $100\%$.
This suggests that \advfnc could be further improved with more data.
Although the attack was carried out using a small dataset, the performance is still comparable to \victimfnc

\subsection{Data Augmentation}\label{data-augmentation}

In order to improve the efficiency of the attack, \adv could craft synthetic queries or choose queries that more effectively explore the decision boundaries.
However, such approaches introduce a pattern to the queries, and there exist techniques that can detect them~\citep{quiring2018forgottensib,juuti2019prada,zheng2019bdpl}.
Instead, \adv can augment their data locally while training \advfnc.
This way, \adv improves the query-efficiency of the attack, without risking detection.

In this section, we study the impact of the following augmentation techniques: horizontal flip, rotation ($5$ degrees), cutout (crop in on a random part of the image), contrast change ($20\%$ increase).
We do not apply the cutout to \superres because the images used during training are already small.
For each task, we train \advfnc with varied amount of data up to the plateau identified in~\Cref{variable-data}: $25\%$, $50\%$, $75\%$ for \mtop, $25\%$, $50\%$ for \stoa, and $25\%$, $50\%$, $75\%$, $100\%$ for \superres.
Each experiment is repeated five times.

\setlength\tabcolsep{4pt}
\begin{table}
    \centering
    \caption{Impact of data augmentation on the performance of \advfnc{s} trained with the subset of attack data that does not reach the plateau: \mtop $50\%$, \stoa $25\%$, and \superres $75\%$ .
    For brevity, we present the results for the comparison with the ground truth (experiment~\ref{exp3}). We report the mean and standard deviation over five runs.}
    \label{tab:augmentation}
    \small
    \begin{tabular}{c|c|c|cccc} \hline
     \multirow{2}{*}{Task}     & \multirow{2}{*}{Test Set} & Baseline          & \multicolumn{4}{c}{Augmentation} \\
                               &                           & (before plateau)  & Flip  & Rotation & Cutout & Contrast \\ \hline
     \multirow{2}{*}{\mtop}    & Monet2Photo Test          & $81.09\pm3.78$    & $76.22\pm3.52$  & $77.84\pm3.6$ & $81.09\pm5.22$ & $80.97\pm3.81$ \\
                               & Landscape                 & $71.42\pm3.01$    & $68.56\pm3.01$ & $67.13\pm2.89$ & $71.42\pm4.67$ & $71.44\pm3.11$ \\ \hline
     \multirow{2}{*}{\stoa}    & Selfie2Anime Test         & $174.42\pm8.89$   & $165.69\pm7.68$ & $170.93\pm8.41$ & $207.04\pm9.13$ & $174.42\pm9.24$ \\
                               & LFW Test                  & $91.42\pm4.04$    & $86.82\pm3.97$ & $89.59\pm4.00$ & $109.20\pm4.07$ & $91.42\pm4.92$ \\ \hline
    \multirow{2}{*}{\superres} & DIV2K Test                & $0.61 \pm 0.08$   & $0.65 \pm 0.07$ & $0.61 \pm 0.07$ & - & $0.61 \pm 0.07$ \\
                               & SRBENCHMARK               & $0.48 \pm 0.01$   & $0.52 \pm 0.01$ & $0.49 \pm 0.01$ & - & $0.48 \pm 0.01$ \\
     \hline
    \end{tabular}
\end{table}

In~\Cref{tab:augmentation} we summarise the impact of data augmentation on the performance of \advfnc.
For \mtop, horizontal flip ($6\%$) and rotation ($4\%$) improve the performance for \advfnc{s} trained with $25\%$ and $50\%$.
These augmentations do not further improve \advfnc trained with $75\%$ of data.
With cutout, the performance does not improve, and the standard deviation increases.
Changing the contrast has no impact on \advfnc trained with $75\%$ of the data.
However, similarly to the cutout, it increases the standard deviation of \advfnc{s} trained with $25\%$ and $50\%$; one of the five \advfnc{s} trained with $25\%$ failed to converge.

For \stoa, horizontal flip ($5\%$) and rotation ($2\%$) improve the performance of \advfnc{s} trained with $25\%$ of the data.
These augmentations do not further improve \advfnc trained with $50\%$.
Cutout hurts the performance of \advfnc by about $15\%$ for \advfnc trained with $50\%$ of data, and by $20\%$ for \advfnc trained with $25\%$.
Similarly to \mtop, changing the contrast has no impact \advfnc trained with $50\%$ of the data.
However, it does increase the standard deviation for \advfnc trained with $25\%$.

For \superres, horizontal flip improves the performance of \advfnc by $4-6\%$ for \advfnc trained with $25\%$, $50\%$, $75\%$ of data; for \advfnc trained with $100\%$ the mean performance does not improve but the standard deviation decreases.
Rotation does not have any impact for any of the experiments.
Changing the contrast, does not affect \advfnc{s} trained with $75\%$ and $100\%$ of the data.
However, it does increase the standard deviation for \advfnc trained with $25\%$ and $50\%$ of the data.

In conclusion, horizontal flip and rotation can improve the performance of \advfnc.
\adv can query fewer samples to achieve the same level of performance.
However, these augmentations do not improve \advfnc beyond the plateau.
On the other hand, cutout and changing the contrast at best do not improve the performance, and at worst are detrimental to it.
We conjecture that changing the contrast does not maintain the target style anymore and hence, leads to worse performance.
Similarly, cutout has minor negative impact on \mtop because the style is consistent across the entire image.
However, for \stoa contextual information about the face might get lost which leads to a major decrease in performance.


\section{Evaluation of Potential Defenses}\label{defense}

We consider several techniques previously proposed to prevent model extraction in other settings, typically in image classification models, to evaluate their applicability to counter model extraction against image translation models.

\subsection{Watermarking}
\noindent\textbf{Trigger-based} watermarking is a class of techniques which embed a watermark into the model by modifying a subset of the training samples (e.g.~\citep{zhang2018protecting, adi2018turning}) or by flipping their corresponding labels~\citep{szyller2020dawn}.

DAWN~\citep{szyller2020dawn} is a watermarking scheme designed specifically to deter model extraction attacks against classifiers.
DAWN changes the prediction label for a small fraction of queries sent by the client and saves them. The query-prediction pairs (trigger set) associated with each client can be then used to verify ownership.
\adv embeds the watermark while training \advfnc with the labels obtained from \victimfnc.
To verify ownership, \victim queries a suspected \advfnc with the trigger set, and declares theft if the predictions match the trigger set.

\begin{figure*}[t]
    \resizebox{1.\columnwidth}!{
    \begin{tabular}{lccr}
        \includegraphics[width=0.25\columnwidth]{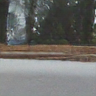} &
        \includegraphics[width=0.25\columnwidth]{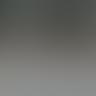} &
        \includegraphics[width=0.25\columnwidth]{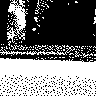} &
        \includegraphics[width=0.25\columnwidth]{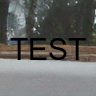} \\
    \end{tabular}}
    \caption{Examples of modifications used to create a trigger set. Left to right: clean sample, Gaussian blur, monochromatic transformation of colours, embedded text.}
    \label{fig:watemarks}
\end{figure*}

Although image translation models do not use any labels, we can adapt DAWN (we use the source code provided by the authors\footnote{\url{https://github.com/ssg-research/dawn-dynamic-adversarial-watermarking-of-neural-networks}}) by combining it with approaches that modify the samples~\citep{zhang2018protecting}.
For a small fraction of queries (0.5\%), we modify the styled image $X_S$, record the pair for future verification, and then return $X_S$ to the client.
For a more detailed explanation of the process, see the original paper~\citep{szyller2020dawn}.
We experimented with three different kinds of modifications: 1) blurring the image; 2) changing colours to monochrome; 3) adding text (see Figure~\ref{fig:watemarks} for examples).

In all three cases, \advfnc learns its primary task and the watermark is not embedded.
In the initial training stages (5-10 epochs), some watermark artifacts are visible.
However as the training progresses, they disappear.

\noindent\textbf{Steganography-based} fingerprinting is an approach where \victim can add an imperceptible bit sequence to all training data, instead of embedding a visible watermark.
We evaluated one such approach which showed that a GAN trained with fingerprinted data will generate images containing that fingerprint~\citep{yu2021artificial}.
To conduct our experiment, we used the source code provided by the authors\footnote{\url{https://github.com/ningyu1991/ArtificialGANFingerprints}}.

\victim jointly trains an autoencoder that embeds the fingerprint, and a detection network that identifies it.
The fingerprint is added to all $X_S$ that are returned to \adv.
However, in our experiments the embedding of the fingerprint fails in one of the two ways: 1) the embedding model collapses and produces random pixels, or 2) the fingerprint is not embedded and the detection accuracy is 50\% (random guess).
We have reached out to the authors but to no avail.

\subsection{Adversarial Examples and Poisoning}

In order to slow down a model extraction attack, \victim can perturb the output of \victimfnc.
Typically, this means adding noise to the prediction logit~\citep{lee2018defending, orekondy2020predictionpoisoning} in order to \emph{poison} \adv's training.
Similarly to watermarking, image translation models do not use any labels.
However, \victim can inject adversarial noise into $X_S$.

We conducted a baseline experiment to determine if this approach is viable.
To each $X_S$, we added a perturbation $\epsilon=0.25$ (under $\ell_\infty$) using projected gradient descent~\citep{madry2017towards}.
The perturbations were crafted with the discriminator of \victimfnc, and using the CleverHans library\footnote{\url{http://www.cleverhans.io}}.
These samples do not transfer to \advfnc's discriminator, and their model trains correctly.

\begin{figure}[t]
    \centering
    \resizebox{1.\columnwidth}!{
    \begin{tabular}{ccc}
        \includegraphics[width=0.33\columnwidth]{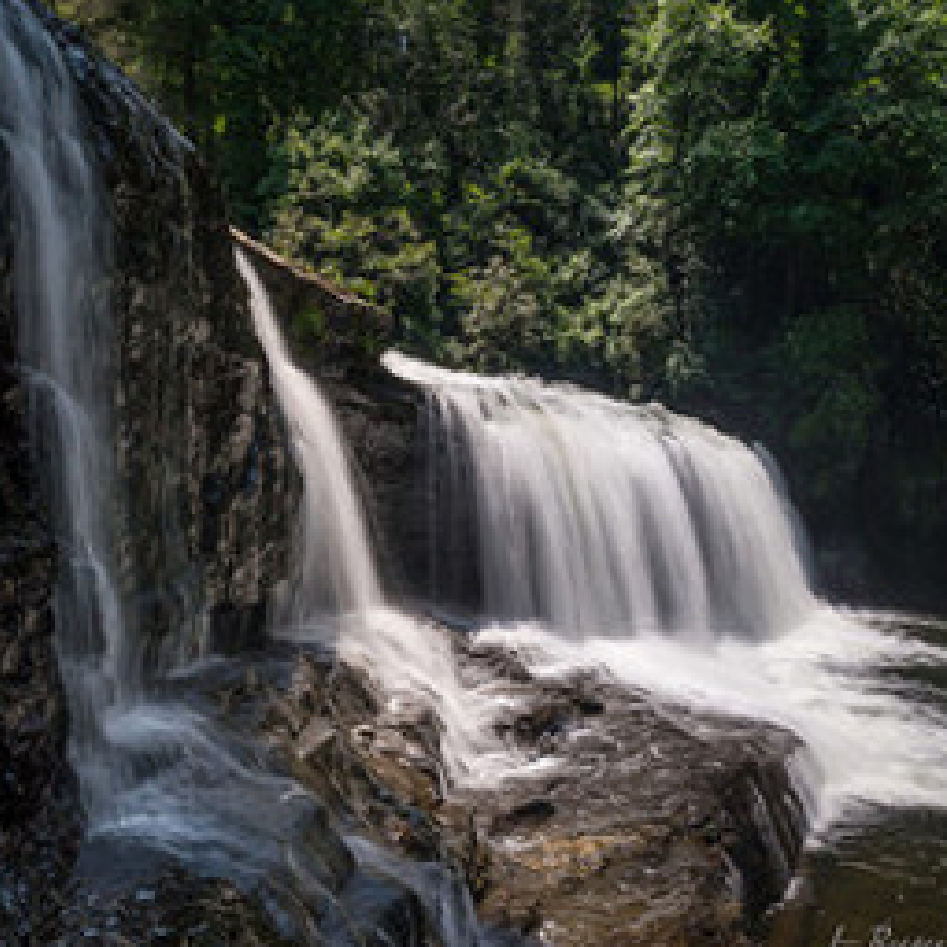} &
        \includegraphics[width=0.33\columnwidth]{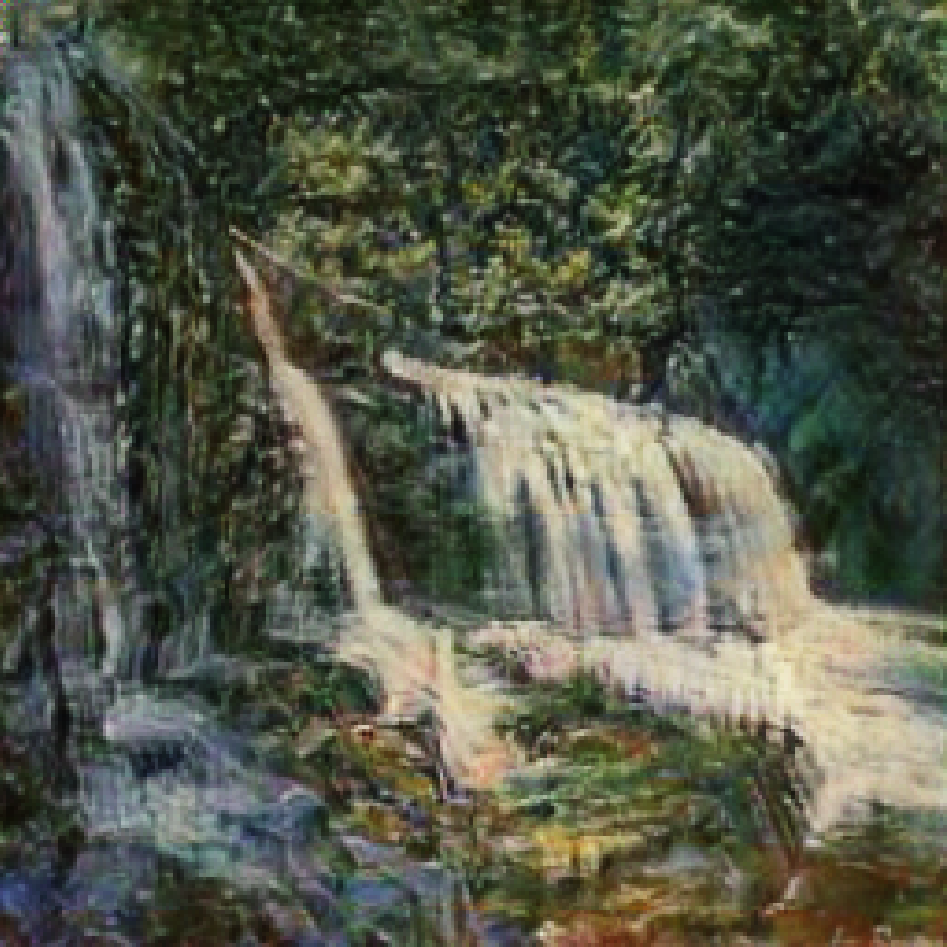} &
        \includegraphics[width=0.33\columnwidth]{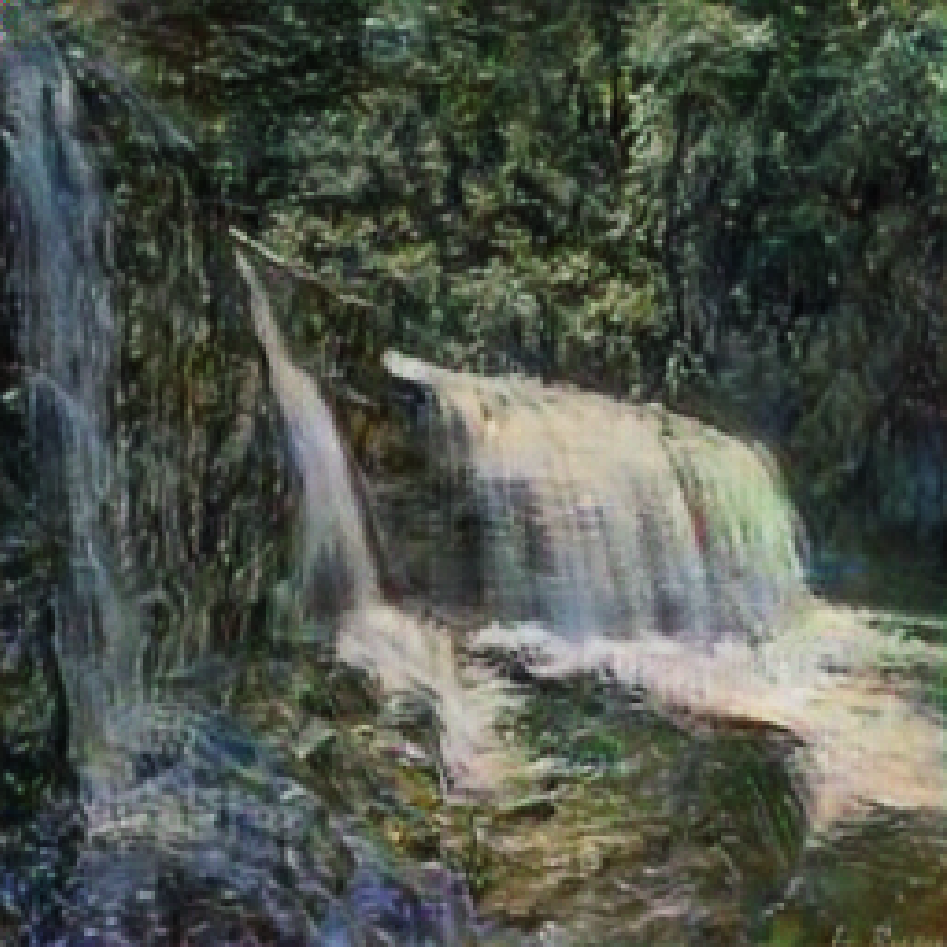} \\
        \includegraphics[width=0.33\columnwidth]{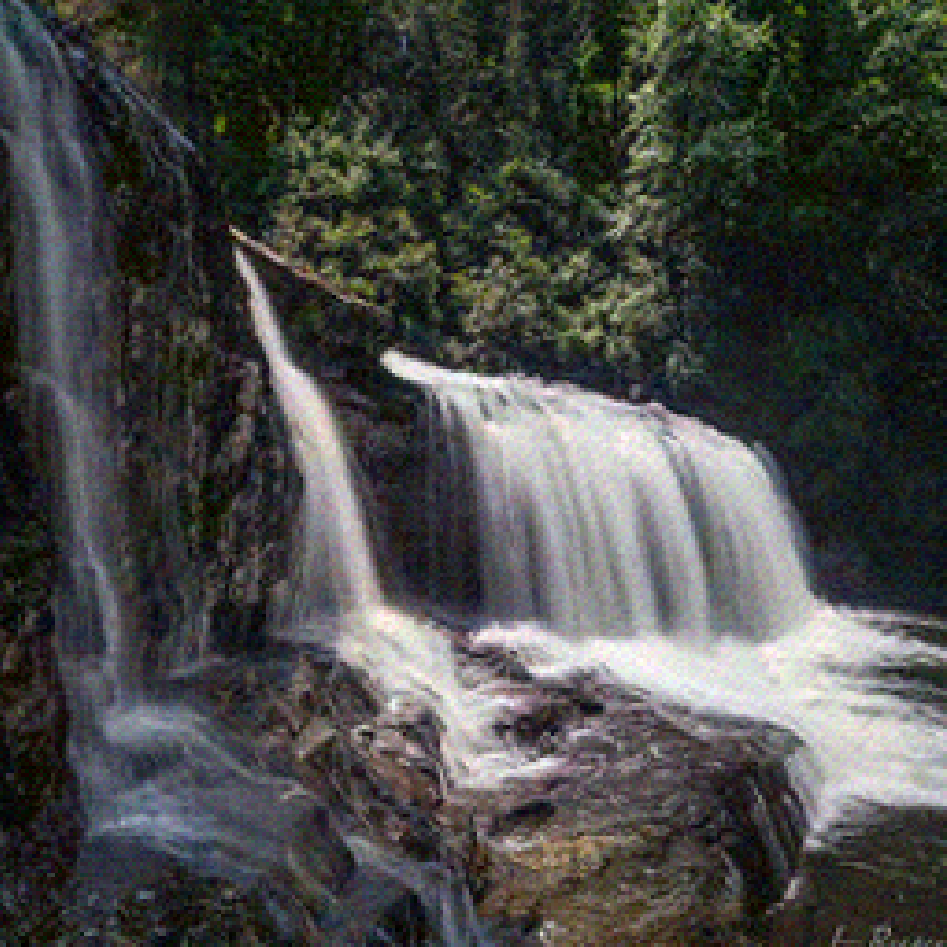} &
        \includegraphics[width=0.33\columnwidth]{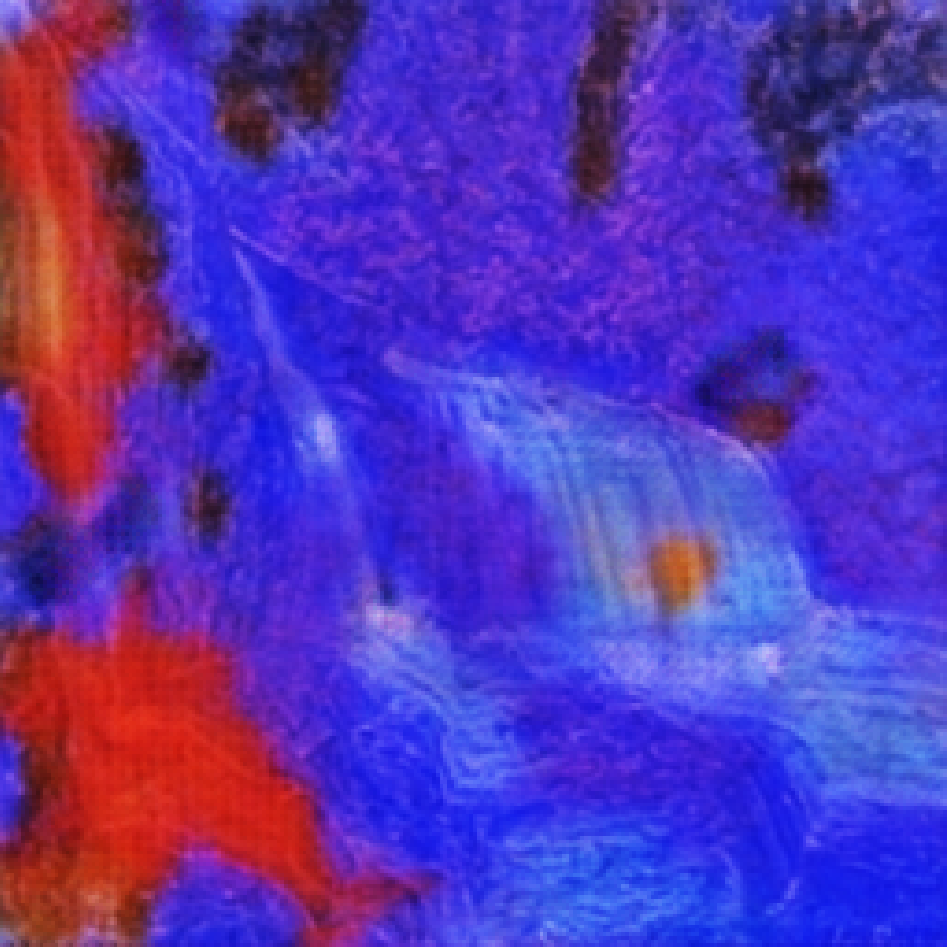} &
        \includegraphics[width=0.33\columnwidth]{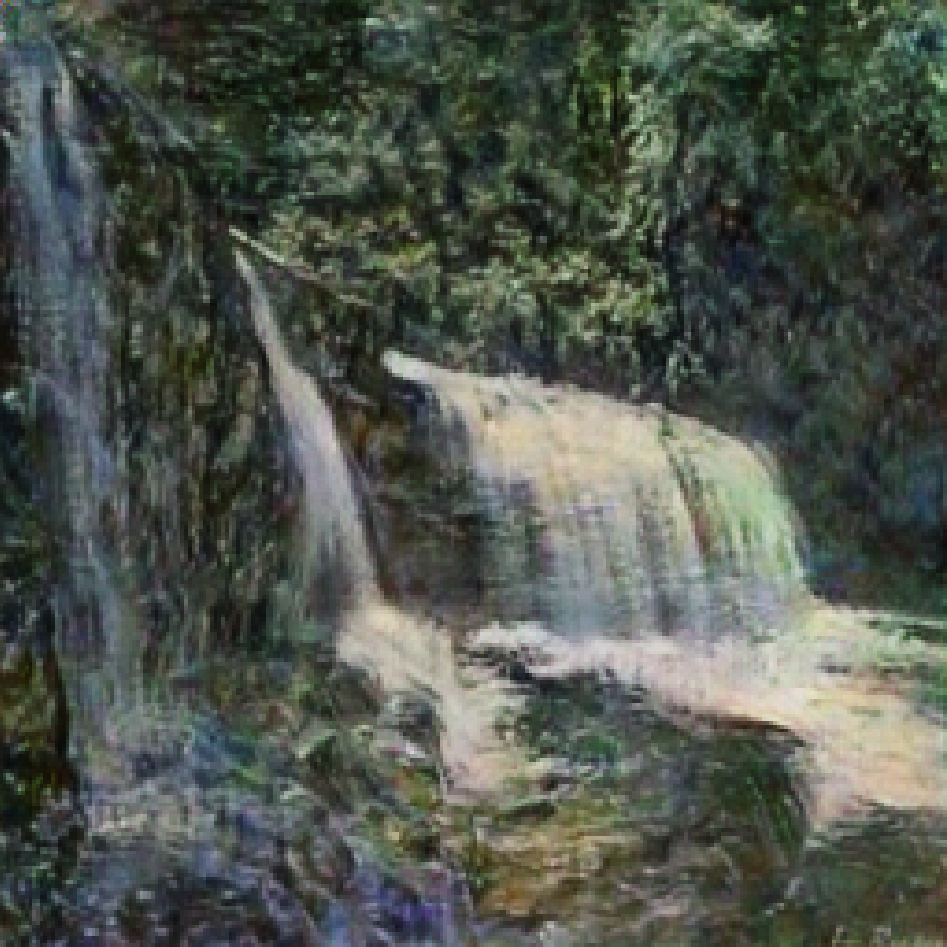} \\
    \end{tabular}}
    \caption{Example translation of an adversarial example crafted using \victimfnc. \textbf{Left}: unstyled, \textbf{center}: \victimfnc output (CycleGAN), \textbf{right}: \advfnc output (Pix2Pix). \textbf{Top row}: using a clean sample as input, \textbf{bottom row}: using an adversarial example as input.}
    \label{fig:adversarial2}
\end{figure}

Furthermore, in our experiments, we discovered that adversarial examples crafted using one generator architecture do not transfer to another one (CycleGAN to Pix2Pix) (\Cref{fig:adversarial2}).

Under realistic assumptions, \victim can craft adversarial examples only using their own models.
Creating an ensemble of GANs, each trained separately for the same style transfer task, in order to improve transferability is too costly.


\section{Discussion}\label{discussion}

\noindent\textbf{Metrics.}
In the \superres setting, we have access to ground truth images in the form of original high resolution photos. Hence, the metrics we used for \superres (PSNR and SSIM) are sufficient for assessing the comparative effectiveness of \victimfnc and \advfnc. This is why a separate user study for \superres was unnecessary.

For the style transfer settings (\mtop and \stoa) there is no such ground truth -- given an input image, there is no quantitative means of assessing if the corresponding output image looks like a Monet paining or an anime figure. Our experience with pixel-based metrics (Section~\ref{metrics}) underscored the possibility that perceptual similarity metrics may not be adequate. This is our motivation for a separate user study to assess the effectiveness in the \mtop and \stoa settings. The distance metric we used (FID) indicated that the output distributions of \victimfnc and \advfnc are close, as did manual inspection of a random subset of the transformations. The study confirmed that both models are similar within reasonable equivalence bounds. However, the user study indicated that \victimfnc and \advfnc are more similar in the \stoa setting than in the \mtop setting (\Cref{fig:study}) whereas the FID metric indicated otherwise (\Cref{tab:models-fid-monet,tab:models-fid-selfie}). This highlights the importance of more robust quantitative metrics for comparing human perception of image similarity.

Additionally, we emphasize that evaluating model extraction attacks against image translation models is more challenging than that of classifiers and NLP models.
In all prior work on the extraction of classifiers, average and per-class accuracy, as well as top-k accuracy, were shown to be sufficiently good for measuring the performance of the surrogate model.

Although surrogate NLP models can be evaluated using e.g. BLEU scores, models with similar scores can produce qualitatively different outputs. Hence, they would benefit from user studies.
Nevertheless, existing extraction attacks against NLP models~\citep{krishna2020thieves,wallace2021translation} did not conduct any user studies.

To the best our knowledge, ours is the first work that uses a user study to evaluate the success of model extraction. We are the first to show that image translation models are also vulnerable, and hence, they need to be protected.

\noindent\textbf{Cost estimate.}
To extract the model, \adv needs to query \victimfnc thousands of times.
It begs the question if the cost of the extraction attack is prohibitive.
We estimate the cost based on the pricing of of OpenAI's models~\footnote{\url{https://openai.com/api/pricing/} Accessed: 2023-01-05.}
At the resolution of $256\times256$, a single query costs \textdollar 0.016.
Therefore, it would cost approximately \textdollar 1250 to extract \stoa model (80,000 queries), \textdollar 160 for \mtop (10,000 queries), and \textdollar 50 for \superres (2,700 queries).

Note that this pricing corresponds to a much bigger and more powerful text-to-image model, and hence, our estimate is an upper bound on the cost of the attack.

\noindent\textbf{Towards a defense.}
In this work, we emphasize the feasibility of attacking realistic \victimfnc, and show that having obtained \advfnc, \adv can launch a competing service.
The next step would be to explore ways of protecting against such attacks.
None of the existing defenses against model extraction attacks explained in Section~\ref{background} apply to image translation models.
Most of these defenses~\citep{juuti2019prada,boogeyman,quiring2018forgottensib,Kesarwani2017model,zheng2019bdpl} rely on examining the distribution of queries from clients to differentiate between queries from legitimate clients and adversaries. In our adversary model, we assumed that \adv uses natural images drawn from the same domain as \victim. As a result, none of those defenses are applicable to our attack. Other defenses~\citep{kariyappa2019adaptivemisinformation,orekondy2020predictionpoisoning} rely on perturbing prediction vectors which is only applicable to attacks against classifiers.

Alternatively, \victim could try to embed a watermark into the model such that all output images contain a trigger that would transfer to \advfnc.
This would allow \victim to prove ownership and deter \adv that wants to launch a competing service but it does not stop the attack on its own.
We adapted four existing watermarking schemes (three designed for image classifiers~\citep{szyller2020dawn, zhang2018protecting}, one for GANs~\citep{yu2021artificial}) to black-box extraction of image translation models (\Cref{defense}).
We show that these schemes do not successfully embed a watermark in \advfnc.
Furthermore, recent work showed that \emph{all} model watermarking schemes are brittle~\citep{lukas2021sok}, and can be removed.

One plausible way to prevent \adv from extracting image translation models is to investigate ways of incorporating adversarial examples and data poisoning into a model as a defense mechanism.
\victim could add imperceptible noise to the output images, designed to make the training of \advfnc impossible or at least slow it down such that it is not economically viable.
However, it was recently shown that such defensive perturbations work only against existing models and are unlikely to transfer to newer and more resilient model architectures~\citep{dixit2021poisoning}.
Hence, such an approach provides only a temporary defense.
Despite this limitation, we evaluated this approach (\Cref{defense}).
In our experiments, adversarial examples crafted using \victimfnc do not transfer to \advfnc.
To overcome this, \victim needs to know the architecture of \advfnc, and craft adversarial examples against it (which \adv could evade by changing the architecture);
or \victim could train several \victimfnc{s} using different architectures, and try to craft examples that transfer.
Because training multiple, different GANs is expensive this approach is not viable.


\noindent\textbf{Ethical considerations.}
Since this work describes attacks, we do not publicly release our code as usual but will make it available only to bonafide researchers, to facilitate reproducibility.
Also, the user study involved human subjects.
However, it was conducted according to our institution's data management guidelines (Section~\ref{userstudy}).

\bibliography{bibliography}
\bibliographystyle{tmlr}

\end{document}